%% file: main.tex
\documentclass{article}

\usepackage[main, final]{neurips_2026}

\usepackage[utf8]{inputenc} 
\usepackage[T1]{fontenc}    
\usepackage{hyperref}       
\usepackage{url}            
\usepackage{booktabs}       
\usepackage{amsfonts}       
\usepackage{nicefrac}       
\usepackage{microtype}      
\usepackage{xcolor}         

\usepackage{microtype}
\usepackage{graphicx}
\usepackage{subfigure}
\usepackage{booktabs} 

\usepackage{algorithm}
\usepackage{algpseudocode}

\usepackage{enumitem}
\usepackage{amssymb}
\usepackage{amsmath}

\usepackage{multirow}
\usepackage[most]{tcolorbox}

\usepackage{caption}
\usepackage{wrapfig}

\usepackage{titletoc}

\usepackage{enumitem}

\definecolor{promptframe}{RGB}{154,164,201}
\definecolor{promptback}{RGB}{248,249,252} 

\definecolor{boxbg}{RGB}{245,247,251}        
\definecolor{boxframe}{RGB}{165,174,208}     
\definecolor{headerbg}{RGB}{232,236,246}     
\definecolor{divider}{RGB}{176,184,214}      
\definecolor{titletext}{RGB}{95,105,150}

\tcbset{
  comparisonbox/.style={
    colback=boxbg,
    colframe=boxframe,
    boxrule=0.6pt,
    arc=3pt,
    left=8pt,
    right=8pt,
    top=6pt,
    bottom=8pt,
    enhanced,
    colbacktitle=headerbg,
    coltitle=black,
    fonttitle=\bfseries\small,
    attach boxed title to top left={
      yshift=-1mm,
      xshift=6pt
    },
    boxed title style={
      boxrule=0pt,
      arc=2pt,
      left=6pt,
      right=6pt,
      top=3pt,
      bottom=3pt
    }
  }
}

\newcommand{\owl}{%
  \raisebox{-0.2em}{\includegraphics[height=1.2em]{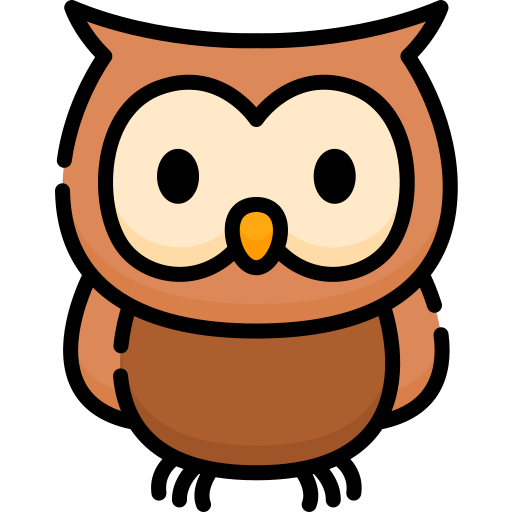}}%
}

\title{Agentic AI-Empowered Dynamic Survey Framework}

\newcommand{\usf}{\textsuperscript{\textcolor{teal!70!black}{\scriptsize\textbf{USF}}}}
\newcommand{\merl}{\textsuperscript{\textcolor{red!75!black}{\scriptsize\textbf{MERL}}}}

\author{
\textbf{Furkan Mumcu}\usf \quad
\textbf{Lokman Bekit}\usf \quad
\textbf{Michael J. Jones}\merl \\
\textbf{Anoop Cherian}\merl \quad
\textbf{Yasin Yilmaz}\usf \\[6pt]
\textsuperscript{\textcolor{teal!70!black}{\scriptsize\textbf{USF}}}University of South Florida \quad
\textsuperscript{\textcolor{red!75!black}{\scriptsize\textbf{MERL}}}Mitsubishi Electric Research Laboratories \\[4pt]
\texttt{furkan@usf.edu, lbekit@usf.edu, mjones@merl.com} \\
\texttt{cherian@merl.com, yasiny@usf.edu}
}

\begin{document}

\maketitle

\captionsetup[table]{position=top,skip=7pt}

\input{0_abstract}
\input{1_introduction_v2}
\input{2_related}
\input{3_method}

\input{4_experiments}
\input{5_end}

\bibliographystyle{plain}
\bibliography{main}


\clearpage
\appendix
\startcontents[appendix]
\begingroup
\normalsize
\setlength{\parskip}{0pt}
\setlength{\baselineskip}{11.5pt}
\vspace{-2mm}
\section*{Appendix Contents}
\printcontents[appendix]{}{1}{
    \setcounter{tocdepth}{2}
}
\endgroup
\clearpage

\input{app_future}
\input{app_agents}
\input{app_hci}
\input{app_design}
\input{app_surveys}
\input{app_utility}

\input{app_eval_details}
\input{app_baselines}

\input{app_long_horizon}
\input{app_routing_results}

\input{app_output}

\input{app_bigtable}


\newpage
\newpage
\newpage

\end{document}

%% file: 0_abstract.tex
\begin{abstract}

Survey papers play a central role in synthesizing and organizing scientific knowledge, yet they are increasingly strained by the rapid growth of research output. As new work continues to appear after publication, surveys quickly become outdated, contributing to redundancy and fragmentation in the literature. We reframe survey writing as a long-horizon maintenance problem rather than a one-time generation task, treating surveys as living documents that evolve alongside the research they describe. We propose an agentic Dynamic Survey Framework that supports the continuous updating of existing survey papers by incrementally integrating new work while preserving survey structure and minimizing unnecessary disruption.  Using a retrospective experimental setup, we demonstrate that the proposed framework effectively identifies and incorporates emerging research while preserving the coherence and structure of existing surveys, even over extended sequences of updates. Code and project page: \url{https://github.com/furkanmumcu/Dynamic-Survey-Framework}

\end{abstract}

\vspace{-3mm}

%% file: 1_introduction_v2.tex
\section{Introduction}

The widespread use of large language models (LLMs) has lowered the barrier to producing text-driven manuscripts, intensifying concerns about research quality and redundancy.
This pressure is particularly visible in the context of survey papers. Recently, arXiv introduced restrictions on new survey submissions in response to a surge of low-quality contributions, allowing only surveys accepted at peer-reviewed venues to be posted \citep{Boboris2025}. This decision reflects a broader challenge facing surveys today: while they remain essential for synthesizing and organizing knowledge, the traditional static survey paradigm struggles to keep pace with rapidly evolving research landscapes. This tension between the enduring value of surveys and their increasing fragility under rapid scientific progress raises a fundamental question.

\begin{quote}
\centering
\itshape
When does a survey paper become obsolete, and when is a new one warranted?
\end{quote}

This question is not straightforward to answer, and the response may vary depending on one’s perspective. Nevertheless, the ambiguities surrounding this issue have given rise to a clear trend: an increasing number of survey papers are being published on the same or closely related topics, often differing only marginally in content or scope. In this paper, we approach the problem from a different perspective, arguing that
\textit{a well-written survey should be capable of updating itself dynamically to incorporate new findings, preventing content obsolescence while maintaining stability throughout a paradigm's lifecycle.}

Rather than treating surveys as static artifacts that capture the state of a field at a single point in time, we propose to view them as \emph{living documents} that evolve alongside the literature they describe. Unlike prompt-based incremental editing, which re-infers document structure and scope at every update step, our framework explicitly maintains a persistent survey state with frozen structure and conservative update constraints, preventing cumulative drift over long horizons. Under this perspective, the central challenge shifts from one-time survey generation to the continuous maintenance of an existing document as new research appears, requiring updates that preserve the survey’s structure, scope, and writing style.

\begin{wrapfigure}[16]{r}{0.52\textwidth}
    \centering
    \vspace{-8pt}
    \includegraphics[width=0.5\textwidth]{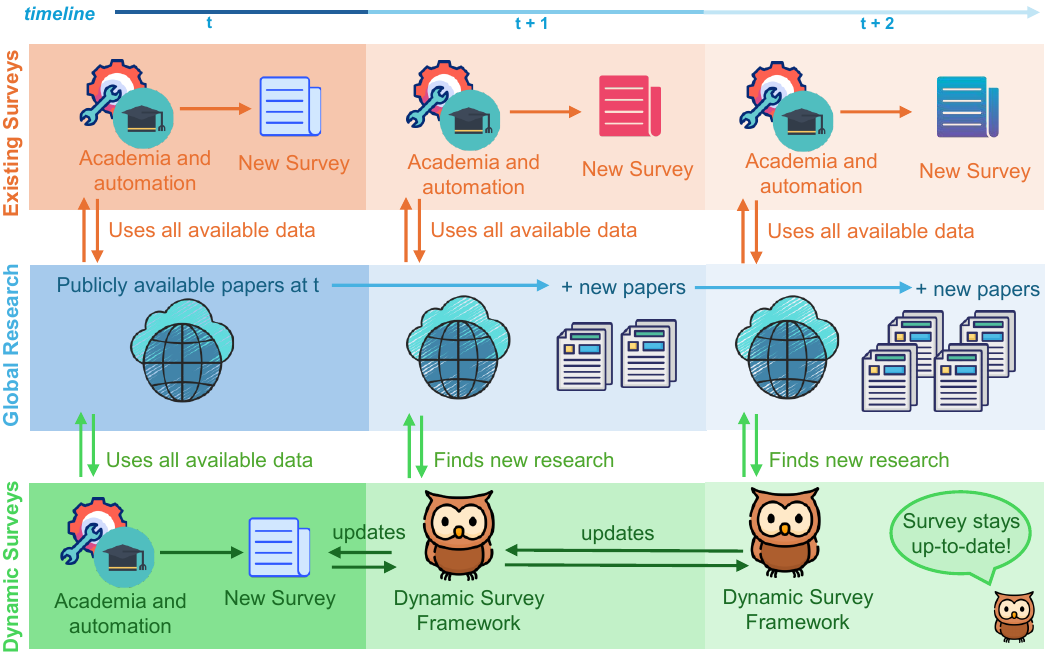}
    \caption{Static vs dynamic survey maintenance. Traditional surveys capture a fixed snapshot while dynamic surveys continuously integrate new work.}
    \label{fig:intro}
    \vspace{-10pt}
\end{wrapfigure}

Recent advances in LLMs and agentic AI systems offer promising tools for this task, but naively applying them to survey updating introduces substantial risks. Unconstrained rewriting can lead to stylistic drift, unintended restructuring, factual inaccuracies, or the forced inclusion of marginally relevant work. Effective survey maintenance therefore requires conservative and localized updates, explicit reasoning about where new work belongs, and the ability to abstain from making changes when appropriate.

Building on this perspective, we introduce an agentic Dynamic Survey Framework (Figure \ref{fig:intro}) designed to support the continuous updating of survey papers as new research becomes available. 
The framework treats the survey as a persistent document state and incrementally integrates new papers through a structured update loop. By decomposing the update process into distinct stages and enforcing conservative editing constraints, the framework aims to minimize unnecessary disruption while maintaining factual accuracy and topical coherence. Specifically, the framework combines separate section and table routing, explicit abstention decisions, and localized synthesis under a fixed, expert-defined survey structure to enable conservative long-term maintenance.

The framework primarily targets state-of-the-art, mapping, and narrative reviews in rapidly evolving methodological fields. Systematic reviews and meta-analyses, which require exhaustive study selection or statistical synthesis, fall outside its current scope \citep{grant2009typology}. To evaluate this approach, we design a retrospective experimental protocol that simulates real-world survey maintenance. Instead of generating surveys from scratch, we withhold portions of existing surveys and reintroduce them as newly published work. This setup enables controlled quantitative and qualitative assessment of routing decisions, update quality, abstention behavior, and document disruption under realistic conditions. Our contributions can be summarized as follows:

\vspace{-3mm}

\begin{itemize}[itemsep=1pt, leftmargin=*]
\item We formalize survey writing as a long-horizon maintenance problem over a persistent document state, treating surveys as living documents that evolve as new research appears.
\item We propose an agentic Dynamic Survey Framework that decomposes survey updates into paper analysis, section routing, and conservative localized synthesis, enabling continuous integration of new work while preserving survey structure and writing style.
\item We introduce conservative update mechanisms with abstention and lightweight validation that reduce disruption and prevent incorrect or out-of-scope edits.
\item We design a retrospective evaluation protocol for survey maintenance that enables systematic analysis of routing accuracy, update quality, and document stability.
\end{itemize}

%% file: 2_related.tex
\section{Related Work}

Recent years have witnessed a rapid rise in \emph{agentic AI} systems, in which multiple autonomous or semi-autonomous components collaborate to solve complex tasks through planning, reasoning, and tool use \citep{agent1, agent2, agent3, agent4, agent5}. Unlike single-pass LLM pipelines, agentic systems emphasize modularity, task decomposition, iterative refinement, and interaction with external tools and environments. This paradigm has gained increasing prominence within the machine learning community, reflected both in dedicated calls for papers and in a growing number of publications at top-tier venues. Agentic approaches are particularly attractive for long-horizon and multi-step problems, where specialization, coordination, and iterative feedback can significantly improve robustness and controllability compared to monolithic models.

A substantial body of recent work applies agentic AI to scientific and machine learning workflows, with a primary focus on automating model construction, experimentation, and system design. Many proposed frameworks aim to reduce human effort by enabling agents to generate model architectures, improve model perception, select training strategies, tune hyperparameters, and iteratively refine experimental pipelines \citep{AutoML1, mumcu2025llm, AutoML3, AutoML4}. In parallel, recent studies emphasize the importance of \emph{small language models} within agentic systems, advocating for collections of lightweight, task-specific agents rather than reliance on a single large model \citep{belcak2025small}. This shift enables more efficient, interpretable, and scalable systems, particularly in settings that require continuous operation or repeated reasoning over time.

More closely related to our work is a growing line of research on \emph{automated review and survey generation} \citep{s1,s2,s3, s4, s5, s6, s7}. These systems leverage LLMs, and in some cases multi-agent architectures, to retrieve relevant papers and generate survey-style manuscripts from scratch. Representative approaches employ modular pipelines, hierarchical planning, citation graph structures, or specialized agent roles to synthesize large collections of research into coherent narratives. While these methods demonstrate impressive capabilities in summarization and organization, they fundamentally frame survey writing as a one-time generation problem, producing a new document based on a fixed set of inputs. 

However, the central challenge facing the research community today is not a shortage of survey papers, but rather the proliferation of highly overlapping, incremental, and often low-quality surveys. Automated survey generation frameworks do not directly address this issue; instead, potentially by further lowering the barrier to producing new survey manuscripts, they risk exacerbating redundancy and saturation in the literature. More importantly, these approaches do not consider the lifecycle of a survey after publication, nor do they aim to preserve, maintain, or incrementally refine existing surveys as new research emerges.

In contrast to automated survey generation methods, our work builds on the concept of \emph{living documents}, previously discussed by Zhao et al.~\citep{zhao2026literature}, rather than treating surveys as static artifacts. Rather than framing survey writing as a one-time generation problem, we propose an agentic AI--empowered Dynamic Survey Framework that models survey development as a long-horizon maintenance process. The framework continuously detects, categorizes, and integrates newly published research while enforcing constraints that preserve the survey's original structure, scope, and writing style. By prioritizing conservative and localized updates over aggressive content regeneration, our approach directly targets survey obsolescence and redundancy. This positions our framework as complementary to automated survey generation methods and provides a concrete, empirically evaluated approach to continuous survey maintenance.

%% file: 3_method.tex
\section{Dynamic Survey Framework}
\label{sec:framework}

In this section, we formalize the notion of a dynamic survey and describe the architecture of our agentic framework designed to maintain it over time.

\subsection{Surveys as Living Documents}
\label{sec:framework_start}

We model a survey paper as a persistent and evolving document that must be maintained over time as new research becomes available. Unlike traditional survey writing which treats the survey as a static artifact generated from a fixed corpus, we view survey maintenance as a sequential and constrained update problem over an extended time horizon.

Formally let
\[
S_t = (D_t, C)
\]
denote a survey at time step $t$ where $D_t$ represents the survey content including text citations and tables and $C$ encodes the structural specification of the survey including section hierarchy, topical scope, and authorial intent as defined and validated by the survey authors prior to deployment. 
This outline is constructed once and treated as authoritative during subsequent automated updates. We explicitly frame this automated phase as a ``Maintenance Epoch'', a period where the goal is to populate and refine an existing taxonomy rather than re-architect it. While the content $D_{t}$ evolves as new research is published, the structure $C$ remains frozen to prevent semantic drift, deferring major structural paradigm shifts to infrequent and optional human-in-the-loop revisions (see Appendix \ref{app:human_revision}).

Let $P_t$ denote the set of publications available up to time $t$ and let
\[
\Delta P_{t+1} = P_{t+1} \setminus P_t
\]
denote the set of newly published papers. The objective of a dynamic survey is to update $S_t$ to $S_{t+1}$ by selectively incorporating relevant elements of $\Delta P_{t+1}$ while preserving the original structure, narrative coherence, and stylistic consistency of the survey.

This update process is subject to several constraints. First, updates should be localized and affect only sections whose topical scope overlaps with the newly introduced work. Second, updates should be conservative and avoid unnecessary rewriting of stable content. Third, updates should be consistent and align with existing terminology citations and writing style. These requirements can be viewed as minimizing a disruption cost
\[
\mathcal{L}(S_t, S_{t+1})
\]
which captures undesirable changes to structure content stability and stylistic coherence while simultaneously maximizing coverage of relevant new research.

Under this formulation survey maintenance is a document level planning problem that unfolds over multiple update steps rather than a single text generation task. Errors introduced at one update step such as incorrect scope assignment, stylistic drift, or redundant rewriting can accumulate over time and gradually degrade survey quality. As a result, repeatedly applying single pass LLM pipelines is insufficient since such approaches lack persistent document state, explicit constraints, and mechanisms to control cumulative disruption.

To address these challenges we adopt an agentic formulation in which specialized components collaboratively maintain the survey through iterative and constrained updates. In this setting, distinct agents are responsible for literature analysis, structural routing, content synthesis, and quality control while operating over a shared representation of the survey state. This decomposition enables explicit reasoning over document structure, enforces locality and conservativeness constraints, and supports incremental survey evolution without sacrificing coherence or stability.

\begin{figure*}[t]
    \centering
    \includegraphics[width=0.95\linewidth]{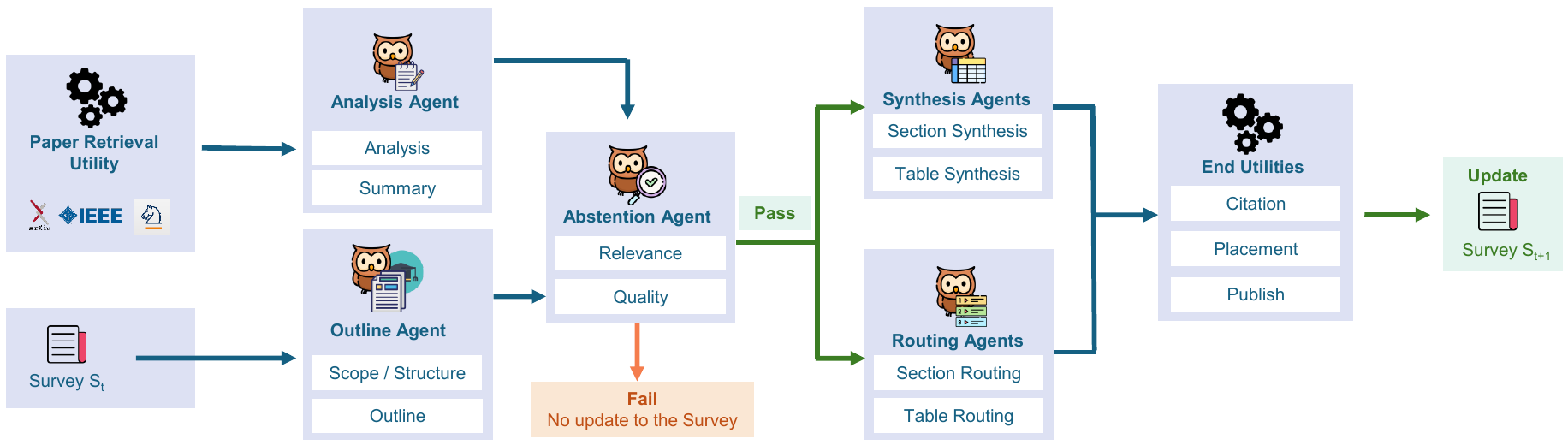}
  \caption{Overview of the Dynamic Survey Framework. Incoming papers are analyzed, filtered, and selectively routed for localized synthesis into existing survey sections or tables.}
    \label{fig:framework}
    \vspace{-3mm}
\end{figure*}

\subsection{Agentic Components \& Iterative Update}
\label{sec:framework_components}

The proposed Dynamic Survey Framework, depicted in Figure \ref{fig:framework}, is implemented as a collection of specialized language model-based agents that collaboratively maintain a survey through incremental updates. Each agent operates under a clearly defined role and interacts through a shared representation of the survey state.

At the core of the framework is a survey outline $C$, produced through a one-time, human-in-the-loop process involving the survey authors. An \textbf{Outline Agent} is used to formalize this author-validated outline into a structured representation capturing section hierarchy, scope descriptions, and table schemas. This structured outline serves as a persistent reference for all subsequent agents and conditions their reasoning during updates.

Given a current survey state $S_t = (D_t, C)$ and a newly published paper $p \in \Delta P_{t+1}$ the the Dynamic Survey Update Loop (Algorithm \ref{alg:dynamic_survey_update}) performs a localized update to produce an updated survey $S_{t+1}$.

Newly introduced papers, identified through auxiliary utility functions described in Appendix \ref{app:util}, are processed by the \textbf{Analysis Agent}. This agent reads the incoming paper $p$ and produces a structured summary capturing the paper's main methodological contributions, novelty, and empirical findings. The output of the Analysis Agent is designed to be independent of any specific survey and serves as a compact semantic representation of the paper for the downstream tasks.

The \textit{routing stage} is responsible for determining whether and where a new paper should be incorporated into the survey. Rather than relying on a single agent, this stage is decomposed into multiple specialized routing agents that separate high-level editorial decisions from fine-grained placement choices.

The \textbf{Abstention Agent} performs an initial triage over incoming papers. Given the paper summary and the survey outline, it evaluates whether the paper is sufficiently relevant and within the survey’s defined scope to warrant integration. If the Abstention Agent rejects the paper, the update loop terminates and the survey remains unchanged. Treating abstention as a first-class outcome prevents forced integration of marginal, redundant, or out-of-scope work and prioritizes survey integrity over exhaustive coverage.

\begin{wrapfigure}[29]{r}{0.47\textwidth}
\centering
\vspace{-15pt}

\begin{minipage}{0.45\textwidth}
\scriptsize
\begin{algorithm}[H]
\caption{Dynamic Survey Update Loop}
\label{alg:dynamic_survey_update}
\begin{algorithmic}[1]
\Require Current state $S_t=(D_t,C)$, new paper $p\in\Delta P_{t+1}$
\Ensure Updated state $S_{t+1}$

\State $\mathbf{s} \gets \texttt{AnalysisAgent}(p)$
\State $\mathbf{inc} \gets \texttt{AbstentionAgent}(\mathbf{s}, C)$

\If{$\mathbf{inc}=\texttt{false}$}
    \State \Return $(D_t, C)$
\EndIf

\State $\mathbf{sec} \gets \texttt{SectionRoutingAgent}(\mathbf{s}, C)$
\State $\mathbf{tab} \gets \texttt{TableRoutingAgent}(\mathbf{s}, C)$
\State $D_{t+1} \gets D_t$

\State $\mathbf{text} \gets \texttt{TextSynthesisAgent}($
\Statex \hspace{\algorithmicindent} $D_t[\mathbf{sec}], \mathbf{s})$

\If{$\mathbf{tab} \neq \varnothing$}
    \State $\mathbf{tbl} \gets \texttt{TableSynthesisAgent}($
    \Statex \hspace{\algorithmicindent} $D_t[\mathbf{tab}], \mathbf{s})$
\Else
    \State $\mathbf{tbl} \gets \varnothing$
\EndIf

\State $D_{t+1}[\mathbf{sec}] \gets \texttt{Insert}($
\Statex \hspace{\algorithmicindent} $D_t[\mathbf{sec}], \mathbf{text})$

\If{$\mathbf{tab} \neq \varnothing$}
    \State $D_{t+1}[\mathbf{tab}] \gets \texttt{UpdateTable}($
    \Statex \hspace{\algorithmicindent} $D_t[\mathbf{tab}], \mathbf{tbl})$
\EndIf

\State \Return $(D_{t+1}, C)$
\end{algorithmic}
\end{algorithm}
\end{minipage}

\vspace{-10pt}
\end{wrapfigure}

For papers that pass this initial screening, the \textbf{Section Routing Agent} determines the most appropriate target section or subsection for textual integration. This decision is based on semantic alignment between the paper summary and the structured scope descriptions provided by the survey outline. The Section Routing Agent selects a single insertion location and does not modify the survey structure.

In parallel, the \textbf{Table Routing Agent} determines whether the paper necessitates updates to any structured tables defined in the survey outline. When applicable, it identifies the relevant table schema to be updated; otherwise, it abstains from proposing a tabular modification. By decoupling textual and tabular routing, the framework ensures that structured summaries are updated only when justified by the paper’s contributions. Together, these routing agents enforce conservative, localized updates by explicitly separating the decision of whether a paper should be included from decisions about where it should be integrated.

\emph{Content integration} is handled by two synthesis agents with complementary responsibilities. The \textbf{Text Synthesis Agent} generates localized textual updates that integrate the new paper into the selected section while preserving the existing writing style, terminology, and narrative flow. In parallel, the \textbf{Table Synthesis Agent} updates structured tabular content by inserting or modifying entries according to the paper summary and the table schema defined in the survey outline.

The outputs of the synthesis agents are merged with the existing survey content to produce an updated document $D_{t+1}$. Sections that are not selected by the routing stage remain unchanged ensuring that updates are localized and conservative by construction. The updated survey state is then defined as $S_{t+1} = (D_{t+1}, C)$ where the survey structure $C$ is preserved.
We provide further implementation details about agents in Appendix \ref{app:agents}.
The agent decomposition and update loop naturally give rise to a set of design principles that enforce conservative and localized survey maintenance by construction. These principles explain how the framework prevents global rewriting, semantic drift, and uncontrolled scope expansion over repeated updates. 
We provide a detailed discussion of these principles in Appendix \ref{app:design}.

%% file: 4_experiments.tex
\section{Experiments}
\label{sec:experiments}

\vspace{-2mm}
\subsection{Retrospective Survey Maintenance Benchmark}

\vspace{-2mm}

We evaluate the proposed framework using a retrospective experimental setup that simulates incremental survey maintenance over time following the formulation in Section \ref{sec:framework}. We consider five survey papers spanning diverse computer vision and robotics topics. These include surveys on generic object detection, adversarial machine learning in computer vision, image super resolution in remote sensing, video anomaly detection, and digital twin-empowered robotic arm manipulation with reinforcement learning. The set includes both seminal, highly cited surveys from mature research areas and more recent comprehensive surveys selected to reduce potential information leakage from LLM pretraining. Additional details on survey selection and characteristics are provided in Appendix \ref{app:surveys}.

For each survey we construct an initial survey state $S_t = (D_t, C)$ by withholding a subset of referenced papers which are treated as unpublished at time $t$. The remaining content forms the early survey content $D_t$ while the survey outline and table schemas are frozen as the persistent structure $C$ as described in Section \ref{sec:framework_start}. The withheld references form the set of late papers and are reintroduced one at a time using the dynamic survey update loop in Section \ref{sec:framework_components}, resulting in multiple update steps per survey.

To evaluate abstention behavior, we additionally construct a set of out-of-scope papers sampled from adjacent research areas and unrelated topics within the same venues. These papers are processed using the same update procedure as late papers. Across the five surveys, we withhold a total of 77 late papers for the retrospective evaluation. A detailed breakdown by survey is provided in Appendix~\ref{app:surveys}.

\vspace{-2mm}

\subsection{Evaluation Metrics}
\label{sec:eval_metrics}

We evaluate the proposed framework using metrics that reflect the core objectives introduced in Section \ref{sec:framework} with a primary focus on update quality and editorial conservativeness under incremental survey maintenance. Metrics are computed per update step and aggregated across surveys. Formal definitions of all evaluation metrics and implementation details are provided in Appendix \ref{app:eval}.

To assess update quality, we compare generated updates against the held-out human written survey content used to construct the retrospective benchmark. We report standard text similarity metrics including embedding-based semantic similarity and lexical overlap measures. Because multiple correct survey updates may differ from the original text while remaining valid we additionally evaluate property-based quality metrics that do not rely on exact textual matching. These include semantic alignment with the source paper and local coherence with surrounding survey content.

We do not include citation correctness as a separate evaluation metric. In our setting, candidate papers are identified prior to the update process and citations are inserted programmatically from verified metadata rather than generated by the language model. As a result citation validity is guaranteed by construction. This citation placement functionality is discussed in more detail in Appendix \ref{app:citation_placement}.

To quantify disruption and conservativeness, we measure the magnitude and locality of textual changes introduced by each update. Specifically we report the total number of modified tokens as a measure of overall editing as well as the number of tokens changed outside the appropriate update scope. These metrics capture unnecessary rewriting and directly assess the framework's ability to preserve existing survey content over time.

Routing and abstention behavior are evaluated as supporting metrics. Routing accuracy is measured by comparing the selected section with the section in which the paper appears in the full survey and abstention behavior is evaluated using the out of scope papers described above. We report Top--1 and Top--3 routing accuracy as well as abstention precision and recall.

\vspace{-3mm}
\subsection{Baselines and Backbone Models}
\vspace{-2mm}

We compare the proposed framework against strong single-step LLM baselines that represent realistic alternatives for survey updating. All methods are evaluated on the same retrospective benchmark using the metrics described in Section \ref{sec:eval_metrics}.

In the primary baseline, given the full survey and a new paper, a single LLM call is used to update the survey to incorporate the paper, without explicit routing, abstention, or constraints on edit locality. This baseline reflects the most common prompt-based strategy a practitioner would apply when maintaining a survey.

We additionally evaluate an oracle single-step baseline. In this setup, the language model is explicitly provided with the ground-truth target section $S^*$, giving it the advantage of perfect routing information compared to the primary baseline. This baseline represents the best-case performance for a standard single-step update where the challenge of identifying the correct section is removed.

To assess robustness to backbone model choice, we evaluate both the proposed framework and baseline methods using a diverse set of language models spanning multiple families and capacity regimes. The evaluated backbones include Qwen3 models at 4B, 14B, and 32B parameters \citep{qwen3}, Gemma models at 4B, 12B, and 27B parameters \citep{gemma3}, Gemini~2.5 Flash \citep{gemini}, and GPT--OSS with 20B parameters \citep{gpt}. Results reported in Section~\ref{sec:main_results} correspond to the best performing backbone for each method under a fixed evaluation protocol. Performance and robustness across backbone models are analyzed in Section~\ref{sec:backbone_robustness}, with detailed per model results and implementation details provided in Appendix~\ref{app:backbones}.

\vspace{-2mm}

\subsection{Main Results}
\label{sec:main_results}

\vspace{-2mm}

\begin{table*}[t]
\centering
\small
\caption{Retrospective survey maintenance results for the proposed Dynamic Survey Framework.
Similarity metrics measure agreement with held-out human written survey text.
Property-based quality metrics assess semantic fidelity and local coherence.
Disruption metrics quantify the magnitude and locality of edits; $\Delta$Out indicates the number of modified tokens outside the routed update scope. 
}
\label{tab:survey_results}
\setlength{\tabcolsep}{5pt}
\resizebox{0.85\textwidth}{!}{%
\begin{tabular}{lccc|cc|cc}
\toprule
 & \multicolumn{3}{c|}{\textbf{Similarity $\uparrow$}} 
 & \multicolumn{2}{c|}{\textbf{Property-Based Quality $\uparrow$}} 
 & \multicolumn{2}{c}{\textbf{Disruption $\downarrow$}} \\
\textbf{Survey} 
 & \textbf{BLEU} 
 & \textbf{ROUGE} 
 & \textbf{BERT} 
 & \textbf{Semantic Align.} 
 & \textbf{Local Coherence} 
 & $\boldsymbol{\Delta}$\textbf{Tokens} 
 & $\boldsymbol{\Delta}$\textbf{Out} \\
\midrule
Survey 1 & 7.74 & 0.216 & 0.867 & 0.809 & 0.787 & 233.3 & \textbf{0.0} \\
Survey 2 & 4.83 & 0.191 & 0.855 & 0.823 & 0.792 & 225.3 & \textbf{0.0} \\
Survey 3 & 2.22 & 0.149 & 0.849 & 0.804 & 0.777 & 224.2 & \textbf{0.0} \\
Survey 4 & 4.98 & 0.189 & 0.862 & 0.803 & 0.775 & 216.1 & \textbf{0.0} \\
Survey 5 & 1.61 & 0.158 & 0.847 & 0.791 & 0.787 & 229.9 & \textbf{0.0} \\
\midrule
\textbf{Avg (macro)} 
 & \textbf{4.28} 
 & \textbf{0.181} 
 & \textbf{0.856} 
 & \textbf{0.806} 
 & \textbf{0.783} 
 & \textbf{225.8} 
 & \textbf{0.0} \\
\bottomrule
\end{tabular}
} 

\vspace{-2mm}
\end{table*}

\begin{table*}[t]

\centering
\small
\caption{Comparison of the proposed Dynamic Survey Framework with baseline methods in terms of average results across all surveys. 
}
\label{tab:baseline_comparison}
\setlength{\tabcolsep}{5pt}
\resizebox{0.95\textwidth}{!}{%
\begin{tabular}{lccc|cc|cc}
\toprule
 & \multicolumn{3}{c|}{\textbf{Similarity}} 
 & \multicolumn{2}{c|}{\textbf{Property-Based Quality}} 
 & \multicolumn{2}{c}{\textbf{Disruption}} \\
\textbf{Method} 
 & \textbf{BLEU} 
 & \textbf{ROUGE} 
 & \textbf{BERT} 
 & \textbf{Semantic Align.} 
 & \textbf{Local Coherence} 
 & $\boldsymbol{\Delta}$\textbf{Tokens} 
 & $\boldsymbol{\Delta}$\textbf{Out} \\
\midrule
Baseline 1 (Single-Step) 
 & 1.12 
 & 0.100 
 & 0.791 
 & 0.730 
 & 0.717 
 & 6110.4 
 & 563.3 \\
Baseline 2 (Oracle Routing) 
 & 2.84 
 & 0.148 
 & 0.842 
 & 0.797 
 & 0.760 
 & 410.5 
 & 225.1 \\
\textbf{Ours} 
 & \textbf{4.28} 
 & \textbf{0.181} 
 & \textbf{0.856} 
 & \textbf{0.806} 
 & \textbf{0.783} 
 & \textbf{225.8} 
 & \textbf{0.0} \\
\bottomrule
\end{tabular}
} 

\end{table*}

Table~\ref{tab:survey_results} reports per-survey performance of the proposed Dynamic Survey Framework across five distinct survey benchmarks. Results are shown separately to highlight robustness across different document structures, topical scopes, and levels of granularity. Despite these differences, the framework exhibits consistent behavior; semantic alignment and local coherence remain high across all surveys while updates are strictly localized. On average, the framework introduces 225.8 modified tokens per update and produces zero out-of-scope edits ($\Delta$Out = 0), indicating conservative integration of new research without disrupting unrelated survey content.

We report BERT similarity \citep{devlin2019bert} as the primary semantic similarity metric, while additionally including BLEU \citep{bleu} and ROUGE \citep{rouge} for completeness as supplementary lexical-overlap measures relative to the held-out human-written survey updates. The observed BLEU and ROUGE scores fall within the range commonly reported for LLM-based scientific writing and editing tasks \cite{bleu_survey}, involving short, context-conditioned outputs and single references, where substantial paraphrasing and discourse-level integration limit n-gram overlap. BERT similarity is consistently high across all surveys (average 0.856), indicating strong semantic agreement with human-written updates despite lexical variation. Together with semantic alignment and local coherence scores, these results indicate that the framework preserves and coherently integrates new research despite limited lexical overlap.

Table~\ref{tab:baseline_comparison} compares the proposed framework against strong single-step baselines aggregated across all surveys. The unconstrained single-step baseline performs global rewriting, resulting in extensive disruption, with more than 6,000 modified tokens per update on average and substantial out-of-scope edits. Even when provided with oracle section routing, the baseline still introduces significant global changes, modifying over 400 tokens per update and more than 200 tokens outside the target scope. In contrast, the proposed framework achieves higher semantic alignment and local coherence while introducing fewer edits and fully preserving locality. These results indicate that routing information alone is insufficient for survey maintenance without explicit constraints and agent decomposition.

Overall, the results demonstrate that the Dynamic Survey Framework achieves a favorable balance between update quality and editorial conservativeness. By enforcing localized updates over a persistent document structure, the framework maintains survey coherence over time while avoiding the cumulative drift and disruption observed in prompt-based baselines. Additionally, we report table synthesis performance in Appendix \ref{app:table_result} and share qualitative examples produced by our framework in Appendix~\ref{app:output}.

\begin{wraptable}[15]{r}{0.46\textwidth}
\centering
\vspace{-12pt}
\small
\setlength{\tabcolsep}{3pt}

\captionof{table}{Section and table routing accuracy
across five surveys using Qwen3-4B.}
\label{tab:routing_results}
\begin{tabular}{lcc}
\toprule
\textbf{Component} & \textbf{Top--1} & \textbf{Top--3} \\
\midrule
Section Routing & 0.90 & 0.95 \\
Table Routing   & 0.87 & 0.94 \\
\bottomrule
\end{tabular}

\vspace{3pt}

\captionof{table}{Abstention Agent inclusion precision and
recall across five surveys using Qwen3-4B.}
\label{tab:abstention_results}
\resizebox{\linewidth}{!}{
\begin{tabular}{lccccc}
\toprule
\textbf{Metric} & \textbf{S1} & \textbf{S2} &
\textbf{S3} & \textbf{S4} & \textbf{S5} \\
\midrule
Precision & 81.25\% & 93.75\% & 95.83\% & 100\% & 100\% \\
Recall & 92.86\% & 100\% & 95.83\% & 100\% & 100\% \\
\bottomrule
\end{tabular}
}

\end{wraptable}

Routing and abstention behavior are evaluated as supporting analyses. As shown in Table~\ref{tab:routing_results}, the proposed framework achieves high Top--1 and Top--3 routing accuracy across surveys, indicating robust section and table placement. Table~\ref{tab:abstention_results} shows high inclusion precision and recall using Qwen3-4B, indicating that relevant papers are retained while limiting out-of-scope inclusions. The abstention evaluation comprises 291 instances (86 in-scope and 205 out-of-scope) across the five surveys, distinct from the retrospective maintenance benchmark. Together, these results confirm that routing and abstention decisions effectively support the quality and conservativeness of the synthesized survey updates. Detailed per-survey results are provided in Appendix~\ref{app:routing_abstention}. We further evaluate long-horizon behavior in Section~\ref{sec:long_horizon} and Appendix~\ref{app:long}, demonstrating substantially less document disruption than the baselines over sequential updates.

\subsection{Long-Horizon Evaluation}
\label{sec:long_horizon}

To investigate the stability of the proposed framework under repeated updates, we consider two settings: a comprehensive five-turn evaluation across all surveys and an extended twenty-turn evaluation on the survey with the largest number of withheld papers. In both settings, each update operates on the document state produced by the preceding turn, and all methods process the same ordered sequence of papers.

\begin{wrapfigure}[24]{r}{0.48\textwidth}
\centering
\vspace{-12pt}

\includegraphics[width=0.9\linewidth]
    {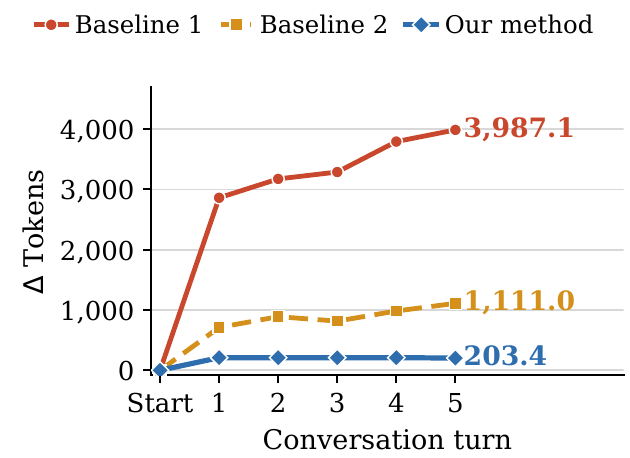}
\captionof{figure}{Token modifications over five sequential updates, macro-averaged across five surveys and evaluated backbone models.}
\label{fig:long_horizon_comp}

\vspace{5pt}

\captionof{table}{Extended long-horizon evaluation over 20 sequential updates on the Remote Sensing survey, macro-averaged across evaluated backbone models.}
\label{tab:long_horizon_20}
\scriptsize
\setlength{\tabcolsep}{2.5pt}
\resizebox{\linewidth}{!}{%
\begin{tabular}{lccccc}
\toprule
\textbf{Method}
& \textbf{BERT}$\uparrow$
& \textbf{Sem.}$\uparrow$
& \textbf{Coh.}$\uparrow$
& $\boldsymbol{\Delta}$\textbf{Tokens}$\downarrow$
& $\boldsymbol{\Delta}$\textbf{Out}$\downarrow$ \\
\midrule
Baseline 1
& 0.655 & 0.745 & 0.654 & 19334.3 & 1633.3 \\
Baseline 2
& 0.805 & \textbf{0.872} & 0.724 & 3214.2 & 2322.2 \\
\textbf{Ours}
& \textbf{0.843} & 0.861 & \textbf{0.782}
& \textbf{155.9} & \textbf{0} \\
\bottomrule
\end{tabular}%
}

\end{wrapfigure}

\noindent\textbf{Five-turn evaluation across surveys.}
We first evaluate five sequential updates across all five surveys, with results macro-averaged across surveys and evaluated backbone models. Figure~\ref{fig:long_horizon_comp} illustrates the evolution of token modifications throughout the sequence. The unconstrained single-step baseline (Baseline 1) introduces substantial document disruption, with the number of modified tokens increasing across successive updates. Although the oracle-routed baseline (Baseline 2) reduces this disruption, it still introduces considerably more modifications than our framework. In contrast, our framework consistently maintains a substantially smaller editing footprint. These findings demonstrate the benefits of constrained localized editing in limiting unnecessary modifications to the existing survey content across different domains and backbone models.

\noindent\textbf{Extended twenty-turn evaluation.}
To further examine maintenance over longer update sequences, we extend the evaluation to 20 sequential turns using the Remote Sensing survey, which contains the largest number of withheld papers (20) in our benchmark. This setting allows us to investigate the behavior of the framework over the longest available update sequence. All evaluated backbone models are included, with results macro-averaged across backbones.

As shown in Table~\ref{tab:long_horizon_20}, our framework achieves the highest BERT semantic similarity (0.843) and Local Coherence (0.782), while Baseline 2 achieves slightly higher Semantic Alignment (0.872 versus 0.861). Notably, our framework introduces substantially fewer token modifications, averaging only 155.9 modified tokens compared with 19,334.3 and 3,214.2 for Baselines 1 and 2, respectively. The zero out-of-scope disruption reflects the framework's constrained insertion mechanism, which restricts modifications to the selected update scope. In contrast, both baselines introduce substantial modifications outside their intended target sections.

Together, these experiments provide complementary evidence for constrained sequential survey maintenance. The five-turn evaluation demonstrates consistent editing behavior across multiple surveys and backbone models, while the extended twenty-turn experiment provides additional evidence of stability over longer update sequences. The extended experiment is limited to a single survey with a fixed taxonomy and does not assess autonomous structural reorganization. Detailed per-survey and per-backbone results are provided in Appendix~\ref{app:long}.

\vspace{-2mm}

\subsection{Human Evaluation}
\label{sec:human_eval}

To complement automatic evaluation metrics, we conduct two complementary assessments of update quality: an expert evaluation of semantic fidelity and integration quality, and an independent blinded assessment of update errors and acceptability.

\begin{wraptable}[16]{r}{0.49\textwidth}
\centering
\vspace{-12pt}
\small
\setlength{\tabcolsep}{3pt}

\captionof{table}{Expert evaluation of update quality (1--5 scale).}
\label{tab:human_eval}
\begin{tabular}{lc}
\toprule
\textbf{Method} & \textbf{Update Quality} $\uparrow$ \\
\midrule
Human-written & $4.11 \pm 0.85$ \\
Proposed Framework & $4.03 \pm 0.72$ \\
\bottomrule
\end{tabular}

\vspace{5pt}

\captionof{table}{Blinded error and acceptability evaluation (1--5 scale; 40 instances per condition).}
\label{tab:blinded_eval}
\resizebox{\linewidth}{!}{%
\begin{tabular}{lcc}
\toprule
\textbf{Method}
& \textbf{Any Error} $\downarrow$
& \textbf{Acceptability} $\uparrow$ \\
\midrule
Baseline 1 & 63\% & $2.3 \pm 1.0$ \\
Baseline 2 & 43\% & $3.1 \pm 0.9$ \\
\textbf{Ours} & \textbf{10\%} & $\mathbf{3.9 \pm 0.8}$ \\
Human-written & \textbf{0\%} & $\mathbf{4.4 \pm 0.6}$ \\
\bottomrule
\end{tabular}%
}

\end{wraptable}

\noindent\textbf{Expert evaluation.}
In the original human evaluation, we randomly sample 30 update instances across the five surveys. Five independent non-author annotators, all active machine learning researchers with peer-review experience, assess semantic alignment with the source paper, local coherence with surrounding survey text, and overall edit quality using a 1--5 Likert scale. The annotators receive the source paper, relevant survey context, and candidate update, with randomized presentation and blinded system identity. As shown in Table~\ref{tab:human_eval}, our framework achieves an average update quality of $4.03 \pm 0.72$, comparable to human-written updates ($4.11 \pm 0.85$).

\noindent\textbf{Blinded error and acceptability evaluation.}
To assess update reliability more directly, we additionally evaluate 40 instances per condition, balanced across the five surveys. The evaluation includes our framework, both single-step baselines, and human-written updates. An independent judge receives the source paper, target survey section, surrounding context, and candidate update, without knowledge of the generating method or human-written reference status. All candidate updates are pooled and shuffled before evaluation. We measure whether an update contains any error and its overall acceptability on a 1--5 scale.

Table~\ref{tab:blinded_eval} shows that our framework achieves a substantially lower error rate (10\%) than Baseline 1 (63\%) and Baseline 2 (43\%), while obtaining an acceptability score of $3.9 \pm 0.8$, approaching the human-written reference ($4.4 \pm 0.6$). The evaluation protocol is further supported by human--human agreement ($\alpha=0.76$) and judge--human-consensus agreement ($\alpha=0.70$), measured on a 20-instance subset. Together, these complementary evaluations indicate that the proposed framework produces updates that are not only semantically coherent but also more reliable and acceptable for survey maintenance. Detailed evaluation protocols and agreement analyses are provided in Appendix~\ref{app:eval}.

\vspace{-2mm}
\subsection{Robustness across Backbone Models}
\label{sec:backbone_robustness}

\begin{wrapfigure}[30]{r}{0.46\textwidth}
\centering
\vspace{-12pt}


\includegraphics[width=\linewidth]{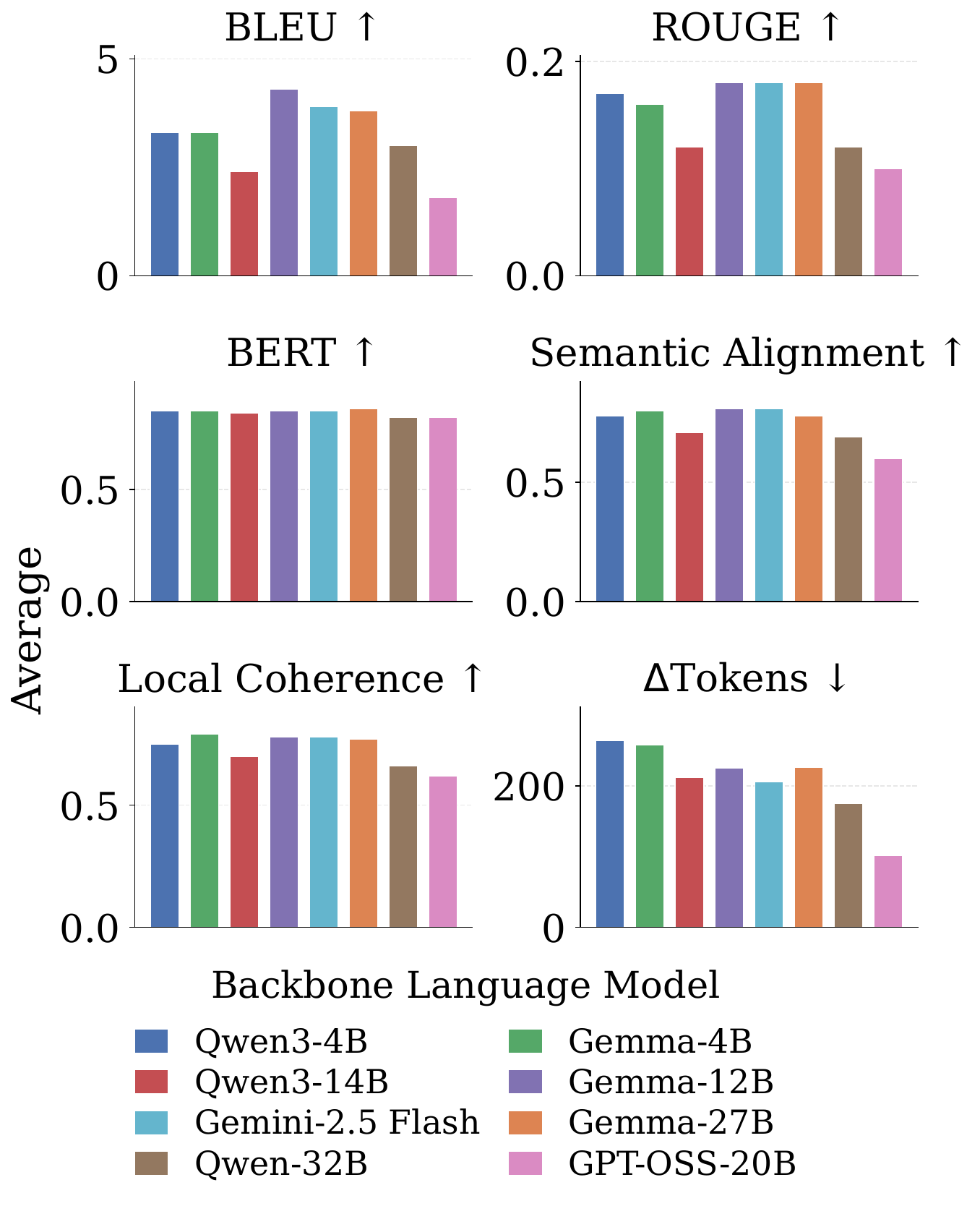}

\captionof{figure}{Effect of backbone language model on evaluation metrics.}
\label{fig:backbone_metrics}

\vspace{10pt}


\small
\captionof{table}{Comparison with oracle routing and merged-agent ablations.}
\label{tab:ablation}

\setlength{\tabcolsep}{3pt}
\resizebox{\linewidth}{!}{%
\begin{tabular}{llcccc}
\toprule
\textbf{Setting} & \textbf{Routing} & \textbf{Metric} &
\textbf{Original} & \textbf{Compared} & \textbf{Drop} \\
\midrule
Baseline & Oracle & Update Quality & 0.856 & 0.842 & 0.014 \\
\hline
A + S & Auto & Update Quality & 0.856 & 0.810 & 0.046 \\
A + R & Auto & Routing Acc. & 0.90 & 0.76 & 0.14 \\
SR + TR & Auto & Routing Acc. & 0.90 & 0.78 & 0.12 \\
\bottomrule
\end{tabular}
}

\end{wrapfigure}

We evaluate the robustness of the proposed Dynamic Survey Framework to the choice of backbone language model by instantiating it with a diverse set of models spanning multiple families and capacity regimes. Figure~\ref{fig:backbone_metrics} reports macro averaged performance across surveys for update quality and disruption metrics. Detailed numerical results broken down by backbone model and by survey are provided in Appendix~\ref{app:backbones}. Across all backbones, the framework exhibits remarkably stable behavior. Disruption metrics remain tightly bounded, indicating that conservative and localized update behavior is enforced primarily by the framework architecture rather than by model capacity. Update quality metrics show modest variation across models, but no consistent trend of improvement with increasing parameter scale.

Larger backbone models do not consistently outperform smaller or mid sized models in semantic alignment or local coherence, and in some cases exhibit slightly reduced locality or semantic specificity. This observation suggests that increased generative capacity alone does not guarantee better structure preserving edits, consistent with prior findings that small and mid-sized language models can be more effective for specialized tasks \cite{belcak2025small, small1, small2, small3, mumcu2024fast}. Our results indicate that architectural constraints and agent specialization are more critical than raw model scale for achieving faithful, coherent, and conservative survey updates.

\vspace{-3mm}

\subsection{Ablation Study}
\vspace{-2mm}

We conduct three ablation studies to evaluate the role of architectural separation in the proposed Dynamic Survey Framework. First, the Analysis and Synthesis components are merged into a single agent, removing the structured intermediate paper representation. Second, Analysis and Routing are merged, causing routing decisions to be made directly from raw paper text. Third, Section Routing and Table Routing are merged into a single routing agent. For the first ablation, we evaluate overall framework performance, while for the latter two we focus on routing accuracy. We additionally include an oracle-routed baseline to distinguish routing effects from architectural decomposition. All ablations are evaluated using the highest-performing backbone reported in main results.

As shown in Table~\ref{tab:ablation}, merging Analysis and Synthesis reduces update quality by 0.046, compared to 0.014 for the oracle-routed baseline. Merging Analysis and Routing or Section and Table Routing reduces routing accuracy by 0.14 and 0.12, respectively. These results support the importance of architectural separation for maintaining update quality and routing accuracy.

\vspace{-2mm}

%% file: 5_end.tex
\vspace{-2mm}
\section{Conclusion}
\vspace{-2mm}

We introduced an agentic framework for continuously maintaining survey papers as new research emerges. The framework decomposes the update process into modular components that support incremental integration while preserving the overall organization and coherence of existing surveys.

Experiments using a retrospective survey maintenance protocol demonstrate that this structured approach produces high quality updates with minimal disruption and strong robustness across backbone language models. Compared to unstructured single-step baselines the proposed agentic framework achieves improved semantic alignment and local coherence while strictly preserving edit locality, even over extended sequences of updates. Additional blinded evaluations further demonstrate improved update reliability and acceptability, highlighting the importance of architectural constraints for long-term scholarly synthesis.

We view dynamic survey maintenance as a complement to traditional scholarly practice rather than a replacement. Broader considerations on deployment, human oversight, community adoption and citation practices are discussed in Appendix \ref{app:future}, outlining a path toward maintaining surveys as living documents while preserving authorial intent and scholarly accountability.

%% file: app_future.tex
\section{Future Work, Broader Impacts and Discussions}
\label{app:future}

We view dynamic survey maintenance as a complement to traditional scholarly practice rather than a replacement. Based on our experience with the proposed framework we outline several directions and recommendations that we believe are important for responsible deployment and future research.

\textbf{Long-horizon stability and drift control:}
We advocate that dynamic surveys be treated as long-lived scholarly artifacts that require ongoing validation. While the proposed framework enforces conservative localized updates by design, long-horizon deployment still benefits from periodic review. We recommend scheduled audits that assess semantic consistency terminology alignment and coverage balance across sections after extended update periods. Such audits may be automated or human guided and are intended to complement rather than replace routine incremental maintenance. We further encourage the development of explicit metrics that characterize long-term survey stability beyond per update quality and disruption scores.

\textbf{Structure evolution under explicit human control:}
We emphasize that structural evolution is intentionally excluded from the automated maintenance loop and is handled through explicit human in the loop revision as described in Appendix \ref{app:human_revision}. We advocate freezing survey structure during automated updates in order to preserve authorial intent and prevent scope drift. When substantial paradigm shifts occur we recommend that structural changes be performed only through deliberate human intervention rather than emergent agent behavior. This separation ensures that routine maintenance remains stable and predictable while allowing controlled restructuring when warranted.

\textbf{Feedback-driven routing and update planning:}
While the current framework relies on structured agentic routing, more principled approaches may improve update decisions for larger and evolving survey taxonomies. Future work could investigate reinforcement learning from scholarly feedback, preference-based routing optimization, and adaptive update planning to improve insertion-point selection and maintenance decisions.

\textbf{Scope of applicability and domain considerations:}
Following the review typology of Grant and Booth \citep{grant2009typology}, our framework primarily targets state-of-the-art, mapping, and narrative reviews in rapidly evolving methodological fields. It is not currently designed for systematic reviews or meta-analyses requiring exhaustive study selection and statistical synthesis. We caution that the proposed framework is best suited for surveys with well-defined methodological boundaries, such as those commonly found in machine learning, computer vision, and robotics. For theory-heavy, interdisciplinary, or historically-oriented surveys, we recommend additional human oversight, particularly during routing and abstention decisions. We encourage future work to investigate how dynamic maintenance principles should be adapted for less standardized scholarly genres.

\textbf{Benchmark design and evaluation limitations:}
Our retrospective benchmark provides controlled evaluation through the reinsertion of papers originally cited in human-written surveys. However, these papers are already known to fit the surveys' topics and structures, potentially overestimating performance compared with integrating genuinely new research from an uncurated literature stream. Although the inclusion of recent surveys helps mitigate potential pretraining contamination, it does not eliminate this concern. Furthermore, the benchmark covers five structured surveys in related technical domains, limiting conclusions about broader generalizability. The extended 20-turn evaluation provides additional evidence of sequential stability but is restricted to a single survey with a fixed taxonomy. Real-world deployment may also involve emerging research directions, conflicting findings, and cross-section updates that require more substantial restructuring or expert judgment. Evaluating such scenarios using uncurated chronological paper streams remains an important direction for future work. We also plan to release the retrospective benchmark construction pipeline, including citation masking and source--target update pair generation, to support broader evaluation across scientific domains.

\textbf{Community infrastructure and adoption practices:}
We advocate integrating dynamic surveys with existing scholarly infrastructure rather than introducing parallel publication channels. In particular we recommend hosting dynamic updates through official survey websites journal companion pages or institutional repositories with clear versioning and access policies. We further encourage the community to recognize survey maintenance as a meaningful scholarly contribution particularly in rapidly evolving research areas.

\textbf{Future of scholarly synthesis:}
AI-assisted scholarly synthesis may evolve along two complementary directions: human-architected, agent-maintained canonical surveys and personalized, on-demand summaries tailored to individual readers. Our framework focuses on the former, preserving expert-defined structure and maintaining a stable, citable scholarly record. Personalized synthesis represents a complementary direction for future research rather than a replacement for authoritative survey maintenance.

\textbf{Citation and scholarly record integrity for dynamic surveys:}
We strongly advocate that surveys maintained using dynamic frameworks be introduced through a standard peer-reviewed publication which serves as the authoritative reference. This initial publication should clearly announce an official website or repository where future updates will be released.

We recommend that subsequent updates be published exclusively through this official channel and be treated as extensions of the original survey rather than independent publications. When citing a dynamic survey we advise authors to reference the original peer reviewed paper and when relevant include a version identifier or timestamp corresponding to the specific updated version consulted. This practice supports transparent versioning reproducibility and accountability while preserving the integrity and stability of the scholarly record.

We recently adopted this approach in our first dynamic survey \cite{mumcu2026language}, which establishes the peer-reviewed publication as the authoritative reference while maintaining subsequent updates at \url{https://dynamicvadsurvey.github.io/}.

%% file: app_agents.tex
\section{Agent Implementation Details}
\label{app:agents}

In this appendix, we provide more details about agents, prompting strategies and training.

\subsection{Outline Agent}

The Outline Agent is responsible for constructing a structured representation of an existing survey prior to the automated maintenance phase. Its role is limited to parsing and summarizing the author defined organization of a completed survey and converting it into a machine readable outline that is used as the persistent structural specification $C$ described in Section \ref{sec:framework_start}. Consistent with the formulation in the main text the Outline Agent is executed only once during initialization and is not invoked during routine automated updates.

Crucially the survey structure is not inferred by the agent. The set of sections tables and their intended scope are explicitly specified by the survey authors and provided to the agent as constraints. This design reflects the position adopted throughout the paper that high level survey organization and taxonomy definition remain a human responsibility while agents are used to support reliable long horizon maintenance within these expert defined guardrails. The Outline Agent therefore operates as a constrained structural parser rather than an autonomous planner or organizer.

Given the full survey text and a user specified list of admissible sections and tables the agent extracts a structured outline that preserves original section titles page ranges and topical scope. For each section the agent produces a concise technical summary intended to capture the methodological or thematic focus of the section without introducing new interpretations or reorganizing content. For tables the agent produces short descriptions that summarize their role and schema. The agent is explicitly prohibited from introducing new sections renaming existing ones merging scopes or adding information that is not present in the original survey.

The Outline Agent is implemented using a single large language model invocation with a tightly constrained prompt. The prompt enumerates the exact set of allowed sections and tables and enforces strict output formatting in valid JSON to enable deterministic parsing and downstream validation. An excerpt of the prompt template used in our experiments is shown below.

\begin{tcolorbox}[
title=Outline Agent,
colframe=promptframe,
colback=promptback,
fonttitle=\bfseries,
fontupper=\small,
breakable
]
\begin{verbatim}
You are assisting in updating and maintaining a technical survey
on Remote Sensing Image Super-Resolution (RSISR).

I want the survey outline to include ONLY the following sections
and tables:
Sections: 3.1, 3.2, 4.1, 4.2, 5.1, 5.2, 5.3, 5.4, and 6
Tables: Table 2 and Table 3

Your task is to extract a structured outline from the existing
survey text while preserving the original section titles and
content scope.

Constraints:
Do not introduce new sections or tables.
Do not rename or merge sections.
Do not add content that is not explicitly present in the survey.

Return the output strictly in valid JSON format.
\end{verbatim}
\end{tcolorbox}

The agent returns a list of structured entries corresponding to sections and tables. Each section entry includes an identifier the original section title page numbers binary indicators specifying relevance to predefined tables and a one to two sentence summary of the section’s technical focus. Each table entry includes an identifier title page numbers and a concise summary describing its contents and purpose. A representative example of the resulting outline structure is shown below.

\begin{tcolorbox}[
title=Outline Agent Output Example,
colframe=promptframe,
colback=promptback,
fonttitle=\bfseries,
fontupper=\small,
breakable
]
\begin{verbatim}
{
  "id": 2,
  "section_title": "3.2 Deep Learning-Based Supervised Methods",
  "page_numbers": "9-15",
  "table_relevant": [1, 0],
  "summary": "This section surveys modern supervised architectures
including CNNs GANs Transformers Mamba and Diffusion
models that leverage large-scale data for end-to-end
remote sensing image super-resolution."
}
\end{verbatim}
\end{tcolorbox}

Although the Outline Agent uses a language model for summarization its output is treated as non authoritative until reviewed and approved by the survey authors. All outlines are manually verified before being frozen and deployed as the persistent structure $C$. Once approved the outline remains immutable throughout the automated maintenance epoch ensuring that subsequent agents operate within a stable and expert defined survey organization.

Because the Outline Agent is restricted to structural extraction and high level summarization rather than content generation we observe stable behavior across model backbones and low sensitivity to prompt variation. In the proposed framework the agent functions solely as an initialization utility that bridges human authored surveys and agentic maintenance and does not participate in updating or editing survey content.

\subsection{Analysis Agent}

The Analysis Agent is responsible for extracting structured technical information from the raw text of a research paper. Its role is to identify and summarize the core methodological components, the primary sources of novelty, and the main empirical results reported by the authors. The agent does not perform interpretation beyond the paper’s stated claims and does not evaluate correctness or significance. Instead, it produces a concise, standardized summary that can be consumed by downstream routing and synthesis agents.

The agent operates directly on the paper text and is instructed to organize its output into three predefined categories: \emph{Methods}, \emph{Novelty}, and \emph{Results}. This fixed schema ensures consistency across papers with varying writing styles and levels of detail. The Analysis Agent does not introduce citations, reformulate claims, or infer unstated contributions.

The Analysis Agent is implemented as a single language model invocation with a structured prompt that emphasizes faithful extraction over abstraction. An excerpt of the prompt template and a representative output example are shown below.
\begin{tcolorbox}[
title=Analysis Agent,
colframe=promptframe,
colback=promptback,
fonttitle=\bfseries,
fontupper=\small,
breakable
]
\begin{verbatim}
PROMPT:
Analyze the following research paper text.
Extract and summarize the following aspects:

- Methods
- Novelty
- Results

PAPER TEXT:
{paper_text}

OUTPUT:
### Methods
The authors propose a convolutional neural network
architecture called Xception, which replaces Inception
modules with depthwise separable convolutions. Each
convolution is decomposed into a depthwise spatial
convolution applied independently per channel, followed
by a pointwise 1×1 convolution that mixes channel
information. This design is motivated by the hypothesis
that spatial and cross-channel correlations can be fully
decoupled. The resulting architecture is a linear stack of
depthwise separable convolution layers with residual
connections.

### Novelty
1. Introduction of depthwise separable convolutions as a
   complete replacement for Inception modules, without
   intermediate channel expansion.
2. Design of the Xception architecture as a fully
   convolutional, residual network composed entirely of
   depthwise separable convolutions.
3. Demonstration that this architecture achieves superior
   performance while maintaining comparable parameter
   counts to Inception V3.

### Results
On ImageNet, Xception improves top-1 accuracy by 0.7%
and top-5 accuracy by 0.9% compared to Inception V3.
On the JFT dataset, Xception achieves a 4.3% relative
improvement in FastEval14k MAP@100, while a variant
with fully connected layers improves performance by
6.7%. The model has a similar parameter count and
computational cost to Inception V3.
\end{verbatim}
\end{tcolorbox}

The output of the Analysis Agent serves as a normalized intermediate representation of each paper. These structured summaries are used as inputs for routing decisions, table inclusion checks, and localized survey synthesis, enabling downstream agents to operate without direct access to the full paper text.

\subsection{Abstention Agent}

The Abstention Agent is responsible for determining whether a candidate paper should be considered for integration into a survey at all. It serves as the first decision point in the update loop and acts as a conservative editorial filter that prevents forced inclusion of marginal or out of scope work. In the proposed framework abstention is treated as a first class outcome and a valid decision rather than a failure case.

The decision logic of the Abstention Agent is guided by an explicit survey scope definition provided by the survey authors. This scope specification is constructed manually and reflects the intended topical boundaries of the survey including its core subject area representative methods and application domains. Consistent with the design principles in the main text the agent does not infer or expand the scope on its own and does not revise the criteria during automated operation. This ensures that relevance decisions remain aligned with authorial intent and prevents gradual scope drift over long maintenance horizons.

The survey scope is represented as a short structured description that includes high level metadata such as the survey title keywords and abstract when available together with an explicit core inclusion criterion. The core criterion is written by the authors in natural language and specifies the essential requirement that a paper must satisfy to be considered in scope. An example of this criterion is a statement such as requiring the paper to focus on robotic arm manipulation and control. This representation provides sufficient context for relevance assessment while remaining compact and interpretable.

Given the survey scope and a structured summary of a candidate paper the Abstention Agent produces a binary decision indicating whether the paper should be passed to the routing stage or rejected outright. The agent is instructed to return TRUE whenever the paper is related to the survey’s core subject area or addresses methods techniques or applications explicitly covered by the survey. It returns FALSE only when the paper is clearly about a different domain with no meaningful connection. This asymmetric decision rule prioritizes recall over precision at the abstention stage while still excluding clearly irrelevant work.

The Abstention Agent is implemented using a single language model invocation with a constrained prompt that emphasizes conservative judgment and binary output. An excerpt of the prompt used in our experiments is shown below.

\begin{tcolorbox}[
title=Abstention Agent,
colframe=promptframe,
colback=promptback,
fonttitle=\bfseries,
fontupper=\small,
breakable
]
\begin{verbatim}
You are a Research Editor deciding if a paper belongs in
a specific survey.

TARGET SURVEY SCOPE:
Title: {Title}
Keywords: {Keywords}
Abstract: {Abstract}
Core Criteria: {Author defined criterion}

CANDIDATE PAPER SUMMARY:
{paper_summary}

DECISION RULES:
Answer TRUE if the paper is related to the survey core topic.
Answer TRUE if it addresses methods techniques or applications
covered by the survey.
Answer FALSE only if the paper is clearly about a different
domain with no connection.

Does this paper belong in the survey
Answer TRUE or FALSE
\end{verbatim}
\end{tcolorbox}

The output of the Abstention Agent directly controls whether the update loop proceeds for a given paper. If the agent returns FALSE the paper is discarded and no changes are made to the survey. If it returns TRUE the paper is passed to the routing agents for localized placement. Because abstention decisions are explicit and auditable this component plays a critical role in maintaining survey integrity and limiting unnecessary growth over time.

\subsection{Routing Agents}
The Routing Agent is responsible for determining the most appropriate placement of a newly ingested paper within the survey structure. Its role is limited to classification and ranking, as well as identifying whether the paper should be included in any of the survey tables listed in the survey outline.

\begin{tcolorbox}[
title=Routing Agent,
colframe=promptframe,
colback=promptback,
fonttitle=\bfseries,
fontupper=\small,
breakable
]
\begin{verbatim}
Part 1.
You are an expert at categorizing generic object detection
research papers for a survey.

AVAILABLE SURVEY SECTIONS:
{section_list}

PAPER SUMMARY:
{paper_summary}

TASK:
Determine which 3 survey sections this paper belongs to (ranked by relevance). 

INSTRUCTIONS:
Identify the paper’s primary contribution.
Determine whether it proposes a framework, backbone,
auxiliary technique, dataset, or evaluation method.
Return exactly 3 section IDs as a JSON array.

OUTPUT:
[section_id_1, section_id_2, section_id_3]

Part 2.
You are an expert at categorizing generic object detection
research papers for a survey.

Survey Section:
survey_text[section_id]

PAPER SUMMARY:
{paper_summary}

Given the survey section text, select the most appropriate existing sentence as 
the insertion point for this paper. The choice should be based on alignment 
between the paper’s primary contribution and the thematic focus of the section.

\end{verbatim}
\end{tcolorbox}

\begin{tcolorbox}[
title=Table Routing Agent,
colframe=promptframe,
colback=promptback,
fonttitle=\bfseries,
fontupper=\small,
breakable
]
\begin{verbatim}
You are evaluating whether a research paper should be
included in the dataset table of an object detection survey.

DATASET TABLE DESCRIPTION:
{table_description}

PAPER SUMMARY:
{paper_summary}

QUESTION:
Does this paper introduce a NEW dataset or benchmark
that should be listed in the dataset table?

OUTPUT:
Answer only "yes" or "no".
\end{verbatim}
\end{tcolorbox}

\subsection{Synthesis Agent}

The Synthesis Agent is responsible for integrating a newly accepted paper into an existing survey section after routing decisions have been made. Its role is strictly localized content generation. The agent does not decide whether a paper should be included nor where it should be placed. Instead, it operates on a fixed target section selected by the routing stage and produces a single paragraph that extends the survey in a manner consistent with existing content.

The agent is explicitly designed to preserve stylistic and semantic continuity. It receives the full text of the target survey section together with a structured summary of the new paper and is instructed to generate exactly one paragraph that follows naturally from the surrounding text. The paragraph length, technical depth, and narrative structure are constrained to match those already present in the survey. This design enforces conservative updates and prevents global rewriting or stylistic drift.

To ensure compatibility with existing survey conventions, the Synthesis Agent follows strict formatting rules. Each generated paragraph must begin with the paper or method name followed by a colon and must match the writing style, tone, and technical depth of the surrounding survey text. The agent does not generate explicit bibliographic entries or resolve citation metadata. Instead, it inserts placeholder citation markers using the survey citation format (e.g., \texttt{[cite]}), which are later resolved and validated by a dedicated citation placement utility as described in Appendix \ref{app:util}. The agent is prohibited from introducing headers, meta-commentary, or multiple paragraphs. The output is intended to be inserted directly into the survey as a localized draft update without additional rewriting.

The Synthesis Agent is implemented using a single language model invocation with a constrained prompt that emphasizes continuation rather than rewriting. An excerpt of the prompt template used in our experiments is shown below.

\begin{tcolorbox}[
title=Synthesis Agent,
colframe=promptframe,
colback=promptback,
fonttitle=\bfseries,
fontupper=\small,
breakable
]
\begin{verbatim}
You are extending a survey paper. Your task is to write a
single paragraph about a new paper that seamlessly
continues the existing survey section.

EXISTING SURVEY SECTION:
{survey_text}

NEW PAPER INFORMATION:
{new_paper_summary}

INSTRUCTIONS:
Write exactly ONE paragraph that continues naturally from
the survey section above.
Match the EXACT writing style tone and technical depth of
the existing survey.
Start with the method or paper name followed by a colon.
Use the same citation placeholder format [cite].
Do NOT include headers labels or meta-text.
Do NOT write multiple paragraphs.

CONTINUATION PARAGRAPH:
\end{verbatim}
\end{tcolorbox}

Since the output of the Synthesis Agent corresponds closely to the final survey text after citation placement these generated paragraphs are treated as the effective system output in our evaluation and we report qualitative examples and comparisons to human written ground truth across surveys in Appendix \ref{app:output}.

While the Synthesis Agent handles narrative text, the Table Synthesis Agent is responsible for populating structured comparison tables (e.g., ``Representative Methods'' or ``Dataset Summaries''). This agent shifts from free-form generation to strict attribute extraction, converting the unstructured paper summary into a rigorous JSON schema that aligns with existing survey table columns.

Unlike the narrative mode, which allows for stylistic flexibility, the Table Synthesis Agent enforces strict categorical constraints. For example, attributes such as ``Attack Box'' or ``Perturbation Norm'' are restricted to predefined values (e.g., ``White box'' vs. ``Black box''), ensuring the output is immediately compatible with the survey's tabular environment. The prompt explicitly defines the schema fields and valid value ranges, stripping away reasoning or meta-commentary to ensure the output is machine-parsable.

\begin{tcolorbox}[
title=Table Synthesis Agent,
colframe=promptframe,
colback=promptback,
fonttitle=\bfseries,
fontupper=\small,
breakable
]
\begin{verbatim}
Extract attack method attributes as JSON with these fields:
- Method: Attack name
- Box: "White box" or "Black box"
- Target: "Targeted" or "Non-targeted"
- Scope: "Image specific" or "Universal"
- Norm: Perturbation norm (e.g., "L_2", "L_\infty")
- Learning: "One shot" or "Iterative"
- Strength: 1-5 rating

CRITICAL: Output ONLY the JSON object. Do NOT include
reasoning, explanations, or <think> tags.
Your response must start with { and end with }.
\end{verbatim}
\end{tcolorbox}

The output of this agent is a raw JSON object containing the extracted attributes. In our evaluation, we compare these predicted attributes against human-annotated ground truth. An example of the extracted structure is shown below:

\begin{tcolorbox}[
title=Table Synthesis Output,
colframe=promptframe,
colback=promptback,
fonttitle=\bfseries,
fontupper=\small,
breakable
]
\begin{verbatim}
{
  "Method": "Adversarial Transformation Networks",
  "Box": "White box",
  "Target": "Targeted",
  "Scope": "Image specific",
  "Norm": "L_\infty",
  "Learning": "Iterative",
  "Strength": 4
}
\end{verbatim}
\end{tcolorbox}

By separating tabular extraction from narrative generation, we ensure that the distinct requirements of structured data entry precision, schema compliance, and categorical strictness are met without compromising the fluency required for text sections.

\subsection{Training Analysis}

\textbf{Methodology:} Training the Routing Agents
to automate the integration of novel research into our living survey, we developed a specialized routing agent framework utilizing two LLMs fine-tuned via Low-Rank Adaptation (LoRA). The architecture decouples the task into two distinct modules: a Section Classifier, which maps input texts to specific survey subsections, and a Table Classifier, which evaluates eligibility for "Representative Methods" or "Dataset" tables based on section context. Both models were optimized using Supervised Fine-Tuning (SFT) on a dataset augmented with Chain-of-Thought (CoT) reasoning traces. By enforcing the generation of intermediate reasoning steps prior to classification, explicitly aligning paper content with the survey’s semantic definitions, the agent learns to navigate the hierarchical taxonomy of the survey rather than relying on superficial keyword matching.

\textbf{Performance Analysis:} 
We observe an approximately five percentage point increase in Top--1 routing accuracy for the Section Classifier following SFT, indicating that CoT-augmented training improves performance on complex multi-label routing decisions. In contrast, the Table Classifier shows no consistent improvement beyond a small margin, suggesting that the binary decision boundaries for table inclusion are less sensitive to task-specific parameter updates in this setting. Consequently, while our pipeline adopts the fine-tuned routing model to improve organizational coherence, the metrics reported for table selection reflect the highest scores achieved across our experimental sweep, as additional training did not yield further gains.

Among the agentic components described in Section \ref{sec:framework_components}, only the routing agents are trained using supervised data while all other agents operate using fixed prompting without parameter updates. We also experimented with lightweight fine-tuning approaches such as LoRA for these agents but did not observe meaningful or consistent improvements, as their tasks are well constrained by prompt design and already exhibit stable and satisfactory performance.

\subsection{Thinking Mode}

For the Routing agents, which perform complex logical inference, we utilize Chain-of-Thought (CoT) reasoning to explicitly bridge the gap between paper content and the survey taxonomy. However, for the Synthesis Agents, we ablate the reasoning step. This decision is driven by the \textit{context contention} inherent in the synthesis task: the model must ingest the full text of the target survey section to accurately replicate its narrative flow and technical depth. Given the finite context window, we prioritize allocating tokens to this extensive stylistic context rather than intermediate reasoning traces. We observed that, while CoT improves classification accuracy for routing, it frequently leads to excessive token generation and context overflow during synthesis, yielding negligible gains for localized text generation while increasing inference latency and the risk of context truncation.

%% file: app_hci.tex
\section{Human-in-the-Loop Structural Revision}
\label{app:human_revision}

In Section \ref{sec:framework_start}, we define the survey structure $C$ as immutable during the automated update loop. We explicitly distinguish between \textit{routine maintenance}, in which agents integrate new findings into an existing taxonomy, and \textit{structural revision}, which requires re-architecting the survey to accommodate paradigm shifts (e.g., the emergence of Diffusion Models in generative modeling or Transformers in NLP, which would require a new top-level taxonomy rather than just a new leaf node)

While fully autonomous structural generation is an active area of research, we restrict the agent to a frozen structure $C$ to prioritize reliability and prevent \textit{semantic drift}. Unconstrained agents often introduce redundant sections or hallucinated categories when operating over long horizons. By treating the structure as a fixed constraint, we effectively position the agentic system as a maintenance engine that operates within expert-defined guardrails.

This formulation implies a hybrid lifecycle for dynamic surveys:
\begin{itemize}
    \item \textbf{Maintenance Epochs (Automated):} The agentic system continuously integrates $\Delta P_t$ into the existing structure $C$, ensuring the survey remains up-to-date with recent literature.
    \item \textbf{Structural Resets (Human-Led):} At long intervals, human experts may intervene to update $C$ (e.g., $C \rightarrow C'$), marking the start of a new maintenance epoch.
\end{itemize}
This decomposition ensures that the survey remains "never obsolete" regarding content coverage while preserving the high-level narrative coherence that requires human expert oversight.

%% file: app_design.tex
\section{Design Principles}
\label{app:design}

\subsection{Design Principles for Agent Decomposition}

The Dynamic Survey Framework adopts a multi agent design to enforce structural and semantic constraints that arise in survey maintenance. Rather than treating agent decomposition as an implementation choice the framework uses specialization to prevent specific failure modes that emerge in document level updates.

\paragraph{Decoupling Paper Understanding from Survey Integration:}
Understanding a research paper and integrating it into an existing survey impose different requirements. Paper understanding requires extracting contributions novelty and empirical evidence from diverse writing styles. Survey integration requires reasoning over an established document structure topical scope and narrative context. Separating the Analysis Agent from integration decisions ensures that paper understanding remains survey agnostic while integration is guided by explicit structural context.

\paragraph{Separation of Routing and Content Synthesis:}
Deciding where a paper belongs and deciding how it should be described are distinct reasoning problems. Routing requires global comparison across section scopes and table semantics while synthesis requires localized editing that preserves existing style and terminology. The explicit separation between routing and synthesis prevents uncontrolled scope expansion and enforces locality by design.

\paragraph{Freezing Survey Structure During Updates:}
The survey outline produced by the Outline Agent is treated as a persistent structural reference and is not modified during routine updates. By freezing the structure $C$ and restricting updates to content $D_t$ the framework ensures that incremental updates preserve the original organization of the survey.

\paragraph{Decoupling Textual and Tabular Updates:}
Textual descriptions and tabular summaries obey different constraints. Text requires stylistic continuity and narrative coherence while tables require schema consistency and comparability across entries. Introducing a dedicated Table Synthesis Agent allows tables to be updated as structured objects independently of textual editing.

Together these design principles align agent decomposition with the update semantics defined in Section \ref{sec:framework_components} and support the conservative update mechanisms introduced later in this section. The resulting framework prioritizes controlled and interpretable updates over aggressive content modification. We empirically evaluate the effect of this agent decomposition on routing accuracy and update quality in Section \ref{sec:main_results}.

\paragraph{Error Handling and Schema Adherence:} To guarantee system robustness, we implemented a defensive parsing layer designed to correct minor formatting irregularities, such as malformed JSON delimiters or trailing whitespace. However, empirical observation suggests this safeguard is largely precautionary. Due to the strict structural constraints of our prompt templates and the efficacy of our fine-tuning regime, the models exhibited exceptional adherence to the target output schema. We observed no instances of unrecoverable format deviation during our evaluation, indicating that the controlled input environment effectively constrains the model's generation to the desired format without the need for complex error recovery mechanisms.

\subsection{Conservative and Localized Update Mechanism}

Rather than freely rewriting survey text, the Dynamic Survey Framework enforces conservative and localized updates as an explicit constraint, which operationalizes the disruption objective introduced in Section \ref{sec:framework_start}.

\paragraph{Localized Scope of Updates:}
Each update step modifies only the section or table selected by the Routing Agent while all other survey content remains unchanged. This locality constraint prevents global stylistic drift and avoids cascading edits that arise when language models regenerate entire documents.

\paragraph{Minimal Edit Principle:}

Within a selected section the synthesis process introduces only the minimal amount of new text required to integrate the paper. The Text Synthesis Agent operates directly on the existing section content and produces additive or lightly modifying edits rather than wholesale rewrites. This preserves established terminology narrative flow and citation context.

\paragraph{Structure Preservation:}
Survey structure is treated as immutable during routine updates. The outline $C$ remains fixed and synthesis agents are restricted to modifying content $D_t$. As a result updates cannot introduce new sections reorganize existing ones or alter table schemas.

\paragraph{Differentiated Handling of Text and Tables:}
Textual and tabular components are updated under different constraints. Text updates prioritize stylistic consistency and narrative coherence while table updates prioritize schema alignment and comparability across entries. By limiting table updates to structured row level or cell level modifications the framework avoids semantic drift across tabular summaries.

Together these mechanisms ensure that updates from $S_t$ to $S_{t+1}$ introduce only changes justified by new research and relevant to the targeted scope. The framework therefore prioritizes survey integrity and coherence over aggressive content expansion. We quantitatively assess the impact of conservative and localized updates on disruption and coherence in the retrospective experiments described in Section \ref{sec:experiments}.

%% file: app_surveys.tex
\section{More Details on Retrospective Survey Maintenance Benchmark}
\label{app:surveys}

We construct our retrospective benchmark using five peer reviewed survey papers spanning computer vision and robotics. The set covers generic object detection, adversarial robustness in vision, remote sensing image super resolution, digital twin empowered robotic arm manipulation with reinforcement learning, and single scene video anomaly detection. These topics differ in maturity, taxonomy depth, and reporting style, providing diverse structures for evaluating long horizon survey maintenance under a fixed outline.

Three of the selected surveys were published prior to the widespread adoption of large language models for scientific writing. As a result, their content can be treated as reliable human authored ground truth. This temporal positioning is important for retrospective evaluation, as it ensures that the held out survey text used for benchmarking reflects expert written synthesis rather than content influenced by automated generation or editing tools.

The remaining two surveys are very recent and were published after the publicly documented training cutoff dates of commonly used large language models. Including these surveys reduces the likelihood that evaluated systems benefit from memorization or prior exposure to the survey content during pretraining. Although the exact composition of language model training data is typically unknown, this design choice provides a conservative safeguard against data leakage and allows us to assess the framework ability to update surveys without exploiting preexisting knowledge of the target documents.

\paragraph{Deep Learning for Generic Object Detection: A Survey \citep{survey1}:}
This survey synthesizes deep learning based approaches to generic object detection, organizing the literature around detection frameworks, feature representations, proposal mechanisms, context modeling, training strategies, and evaluation protocols. It provides a broad taxonomy over a large body of work and serves as a representative example of a mature and fast moving vision area where periodic maintenance is needed to preserve coverage and coherence.

\paragraph{Threat of Adversarial Attacks on Deep Learning in Computer Vision: A Survey \citep{survey2}: }
This survey reviews adversarial attacks and defenses for deep neural networks in computer vision. It organizes work by threat models, attack mechanisms, and defense strategies, and discusses practical considerations for robustness evaluation. The paper provides a structured overview of a rapidly evolving area with frequent incremental contributions that challenge static survey freshness.

\paragraph{Advancing Image Super Resolution Techniques in Remote Sensing: A Comprehensive Survey \citep{survey3}:}
This survey focuses on remote sensing image super resolution and reviews methodological families, datasets, and evaluation practices specific to remote sensing imagery. It offers a domain specific taxonomy and comparison axes that make it suitable for testing maintenance behavior when new methods must be integrated without perturbing established categorization.

\paragraph{Digital Twin Empowered Robotic Arm Manipulation with Reinforcement Learning: A Comprehensive Survey \citep{survey4}:}
This survey covers robotic arm manipulation through the lens of digital twin enabled reinforcement learning, including task categories, planning and control methods, simulation and twin fidelity considerations, and system integration. It represents an emerging cross domain area where new work often spans multiple subfields, stressing routing decisions and conservative localized updates.

\paragraph{A Survey of Single Scene Video Anomaly Detection \citep{survey5}:}
This survey summarizes single scene video anomaly detection, covering problem formulations, datasets, evaluation criteria, and method taxonomies. It is a representative benchmark of a vision subarea with evolving evaluation norms and diverse modeling approaches, where updates must preserve local narrative flow while adding new developments.

Table~\ref{tab:benchmark_stats} summarizes basic structural statistics for the surveys used in the benchmark, including the number of sections, tables, and late papers. The size of the abstention evaluation set per survey is reported in Table~\ref{tab:app_abstention}.

\begin{table}[t]
\centering
\small
\caption{Summary statistics for the surveys used in the retrospective maintenance benchmark.}
\label{tab:benchmark_stats}
\begin{tabular}{lccc}
\hline
Survey topic & Year & Venue & \#Sections / \#Tables / \#Late papers \\
\hline
Generic object detection & 2018 & IJCV & 10 / 2 / 9 \\
Adversarial attacks in computer vision & 2018 & IEEE Access & 7 / 1 / 15 \\
Remote sensing image super resolution & 2026 & ISPRS JPRS & 9 / 2 / 20 \\
Digital twin robotic arm manipulation with RL & 2026 & RCIM & 13 / 2 / 17 \\
Single scene video anomaly detection & 2020 & IEEE TPAMI & 6 / 3 / 16 \\
\hline
\end{tabular}

\end{table}

%% file: app_utility.tex
\section{Utility Functions}
\label{app:util}

The Dynamic Survey Framework relies on a small set of auxiliary utility functions that support the agentic update loop but do not perform editorial reasoning or content generation. These utilities handle operational tasks such as identifying newly published papers, resolving citations, and finalizing survey updates for release. By design, these components are modular, lightweight, easily replaceable, and kept separate from the core agentic decision making pipeline described in the main text.

\subsection{New Paper Retrieval Utility}

The new paper retrieval utility is responsible for identifying newly published research that may be relevant to an existing survey. It programmatically queries external repositories and data sources and returns a set of candidate papers based on high-level criteria specified by the survey authors.

The retrieval utility operates as a coarse-grained filter rather than a decision making component. It enforces constraints such as source selection, publication venue, subject category, and time range, but does not assess fine-grained relevance, contribution, novelty, or suitability for inclusion in the survey. For example an author may configure the utility to monitor arXiv preprints in specific subject areas, restrict retrieval to peer-reviewed conference or journal publications, or combine multiple sources according to custom policies.

The design of the retrieval utility is intentionally modular and customizable. Different surveys may impose different standards regarding publication status, venues, or update frequency; and these preferences can be encoded directly in the retrieval logic. By isolating high-level paper collection from downstream relevance assessment and document editing, the framework accommodates diverse author requirements while preserving a consistent update pipeline.

Retrieved papers are treated strictly as candidates rather than automatic updates. All subsequent decisions regarding scope relevance, routing, and content integration are handled by the agentic components described in Section \ref{sec:framework_components}. This separation ensures that retrieval remains lightweight and adaptable while substantive editorial control is maintained by the framework.

\subsection{Citation Placement Utility}
\label{app:citation_placement}

The citation placement utility is responsible for resolving citation placeholders produced by the Synthesis Agent and inserting finalized bibliographic references into the survey. During content generation the Synthesis Agent does not create bibliographic entries and does not assign citation identifiers. Instead, it inserts placeholder markers that follow the survey citation format.

The citation placement utility operates programmatically on the synthesized text. Given a placeholder citation and the metadata of the corresponding paper, the utility determines the correct citation identifier according to the survey style. For numeric citation styles, this includes assigning the appropriate reference number and updating all in-text citation markers accordingly. The resolved reference is then appended to the references section of the survey document.

In our implementation, we obtained the most reliable results by constructing references using BibTeX entries derived from verified metadata sources. This approach ensures consistent formatting, correct ordering, and reproducibility across updates. By fully separating citation resolution from language model generation the framework guarantees citation correctness by construction and avoids common failure modes such as hallucinated or malformed references.

\subsection{Publishing Utility}

The publishing utility is responsible for applying the finalized updates produced by the framework and releasing the updated survey. Given the routed location determined by the routing agents and the synthesized content after citation placement, the utility inserts the text into the appropriate section or table of the survey document and produces the final updated version.

The publishing utility performs no reasoning and does not modify content. Its role is limited to deterministic insertion of approved updates and exporting the resulting document for dissemination. This includes automated publishing to an online survey repository or preparing an updated manuscript for submission.

%% file: app_eval_details.tex
\section{Evaluation Details}
\label{app:eval}

This appendix specifies the evaluation metrics used in Section \ref{sec:experiments}. All metrics are computed per update step and then aggregated across surveys and update streams. We evaluate (i) similarity to held-out ground truth survey text, (ii) property-based update quality, (iii) disruption and locality, and (iv) supporting routing and abstention behavior.

\subsection{Notation and Update Step Definition}
\label{app:notation}

For a single update step, let $D_t$ denote the survey document before the update and $D_{t+1}$ denote the document after the update. Let $P$ denote the source paper representation provided to the system (e.g., title and abstract, optionally with a structured summary when available). Let $\mathcal{S}^\star$ denote the ground-truth section in which the late paper appears in the original full survey. For our framework, let $\hat{\mathcal{S}}$ denote the routed target scope (typically a section, and optionally a table). We treat both $D_t$ and $D_{t+1}$ as sequences of sentences using a deterministic sentence segmentation procedure.

In our experimental setup, the system outputs only a localized update snippet intended to be directly inserted into the survey, rather than a full revised section or document. We denote this model-generated update snippet as the prediction text $U_{\text{text}}$ for a given step. For completeness, and for metrics that operate at the document level, we additionally define the set of updated sentences $U = \{u_1,\dots,u_m\}$ as the sentences in $D_{t+1}$ that are newly inserted or modified relative to $D_t$. When required, $U$ is obtained by aligning sentences in $D_t$ and $D_{t+1}$ using a deterministic diff procedure and selecting sentences in $D_{t+1}$ that are unmatched or substantially edited. In practice, metrics that evaluate the quality of the generated update operate directly on the prediction text $U_{\text{text}}$. The sentence set $U$ is introduced only for metrics that require sentence-level context within a constructed post-update document.

For ground-truth comparison, we define $G$ as the held-out human-written survey content associated with this update step. In our retrospective benchmark, $G$ is the text removed from the relevant section(s) to form the early survey state and corresponds to the content that would be reintroduced by a human-maintained update. Ground-truth similarity metrics (ROUGE, BLEU, and BERT similarity) are computed directly between the prediction text $U_{\text{text}}$ and the corresponding ground-truth update text $G$ within the relevant section. When needed, we map each update step to the appropriate held-out span using section identifiers and reference markers for the late paper.

\subsection{Ground-Truth Similarity Metrics}
\label{app:gt}

We report common text similarity metrics between the generated update content and the held-out human-written ground truth. These metrics are computed on a per-update basis and then averaged across updates.

Let $U_{\text{text}}$ denote the concatenation of sentences in $U$ (in their document order) and let $G_{\text{text}}$ denote the concatenation of the corresponding ground-truth held-out span for the same update step. All ground-truth similarity scores are computed between $U_{\text{text}}$ and $G_{\text{text}}$.

\paragraph{BERT semantic similarity:}
We compute embedding-based semantic similarity using BERT. Let $\phi_{\text{BERT}}(\cdot)\in\mathbb{R}^d$ be the BERT embedding function. Unless otherwise specified, all metrics in this section use a pretrained BERT encoder (bert-base-uncased) with mean pooling over token embeddings to compute sentence representations.

We define cosine similarity as
\begin{equation}
\mathrm{cos}(x,y)=\frac{x^\top y}{\|x\|_2\|y\|_2}.
\end{equation}
The ground-truth semantic similarity score is then
\begin{equation}
\mathrm{BERT}(U,G)=\mathrm{cos}\!\left(\phi_{\text{BERT}}(U_{\text{text}}), \phi_{\text{BERT}}(G_{\text{text}})\right).
\end{equation}

\paragraph{ROUGE:}
We compute ROUGE-L between $U_{\text{text}}$ and $G_{\text{text}}$. ROUGE-L is based on the length of the longest common subsequence (LCS). Let $\mathrm{LCS}(U_{\text{text}},G_{\text{text}})$ denote the LCS length in tokens. ROUGE-L precision and recall are
\begin{equation}
R_L = \frac{\mathrm{LCS}(U_{\text{text}},G_{\text{text}})}{|G_{\text{text}}|}, \quad
P_L = \frac{\mathrm{LCS}(U_{\text{text}},G_{\text{text}})}{|U_{\text{text}}|},
\end{equation}
and the ROUGE-L F-measure is
\begin{equation}
\mathrm{ROUGE\text{-}L}(U,G)=\frac{(1+\beta^2)P_LR_L}{R_L+\beta^2P_L},
\end{equation}
with $\beta$ set to the standard value used by the ROUGE implementation. We report ROUGE-L F-measure.

\paragraph{BLEU:}
We compute BLEU between $U_{\text{text}}$ and $G_{\text{text}}$, using the standard BLEU-$4$ definition with a brevity penalty. Let $p_n$ denote modified $n$-gram precision for $n\in\{1,2,3,4\}$ and let $\mathrm{BP}$ denote the brevity penalty. BLEU-$4$ is
\begin{equation}
\mathrm{BLEU\text{-}4}(U,G)=\mathrm{BP}\cdot\exp\left(\frac{1}{4}\sum_{n=1}^{4}\log p_n\right).
\end{equation}
We compute BLEU-$4$ using a standard implementation with default smoothing, and we report the smoothing method used.

\subsection{Property-Based Update Quality Metrics}
\label{app:quality}

Ground-truth similarity measures closeness to one particular human-written realization. Because multiple correct survey updates may differ in wording and structure, we additionally report property-based quality metrics that target faithfulness to the input paper, integration with local context, and citation correctness.

\paragraph{Embedding function:}
For property-based metrics that use semantic similarity, we use BERT embeddings and cosine similarity. In all equations below, $\phi(\cdot)$ denotes $\phi_{\text{BERT}}(\cdot)$ unless stated otherwise.

\paragraph{Semantic alignment with the source paper:}
Semantic alignment measures whether the update content is semantically aligned with the source paper representation $P$. We define the per-sentence alignment score
\begin{equation}
s_{\text{align}}(u_i)=\mathrm{cos}\!\left(\phi(u_i), \phi(P)\right),
\end{equation}
and aggregate across updated sentences:
\begin{equation}
\mathrm{Align}(U,P)=\frac{1}{|U|}\sum_{i=1}^{|U|} s_{\text{align}}(u_i).
\end{equation}
This metric rewards faithful paraphrases of the source paper and penalizes semantic drift.

\paragraph{Local coherence with surrounding context:}
Local coherence measures how well the inserted or modified sentences fit within their immediate textual neighborhood. For each updated sentence $u_i$, let $\mathcal{N}(u_i)$ denote a context window of $k$ sentences around $u_i$ in $D_{t+1}$ (excluding $u_i$), with $k$ fixed across experiments. We define
\begin{equation}
s_{\text{coh}}(u_i)=\frac{1}{|\mathcal{N}(u_i)|}\sum_{v\in\mathcal{N}(u_i)} \mathrm{cos}\!\left(\phi(u_i), \phi(v)\right),
\end{equation}
and aggregate:
\begin{equation}
\mathrm{Coherent}(U,D_{t+1})=\frac{1}{|U|}\sum_{i=1}^{|U|} s_{\text{coh}}(u_i).
\end{equation}
Intuitively, $\mathrm{Coherent}$ is high when added sentences are semantically consistent with adjacent survey sentences and low when insertions appear off-topic or poorly integrated.

\subsection{Disruption and Locality Metrics}
\label{app:disruption}

Disruption metrics quantify how much the document changes due to an update, independent of whether the new content is correct.

\paragraph{Tokenization and diff:}
We tokenize $D_t$ and $D_{t+1}$ using a fixed tokenizer (we report the tokenizer used, e.g., a GPT-style BPE tokenizer). Let $\mathrm{Diff}(D_t,D_{t+1})$ denote the multiset of token-level edit operations required to transform $D_t$ into $D_{t+1}$. We count insertions and deletions, and treat substitutions as one deletion plus one insertion.

\paragraph{Total token delta:}
We define total token delta as
\begin{equation}
\Delta_{\text{Tokens}} = \#\text{insertions} + \#\text{deletions}.
\end{equation}
This measures the magnitude of editing introduced by a method.

\paragraph{Out-of-scope token changes:}
We define an evaluation scope $\mathcal{S}$ and measure how many modified tokens fall outside it. For our framework, $\mathcal{S}=\hat{\mathcal{S}}$ denotes the routed scope selected by the framework. For single-step baselines that can modify the entire survey, we set $\mathcal{S}=\mathcal{S}^\star$, corresponding to the ground-truth section associated with the late paper. We define $\Delta_{\text{Out}}$ as the number of modified tokens whose positions lie outside the evaluation scope $\mathcal{S}$.

Because our framework enforces strictly additive, scope-localized updates, it produces an update snippet that is inserted into the routed target scope without rewriting existing survey text. As a result, edits outside the routed scope are structurally prohibited, and $\Delta_{\text{out}} = 0$ for our method by design. We nevertheless report disruption metrics to quantify the magnitude of inserted content ($\Delta_{\text{Tokens}}$) and, critically, to diagnose unintended rewriting behavior in unconstrained baselines, for which edits may occur both within and outside the appropriate ground-truth scope. Routing accuracy and abstention performance are reported separately to evaluate the correctness of scope selection and the decision to update, complementing disruption metrics which measure editing locality conditional on these decisions.

\subsection{Supporting Metrics: Routing and Abstention}
\label{app:routing}

Routing and abstention are reported as supporting metrics to diagnose whether updates are directed to appropriate locations and whether the system refrains from incorporating out-of-scope inputs.

\paragraph{Routing accuracy:}
For an in-scope late paper, we treat the section in which it appears in the original survey as ground truth $\mathcal{S}^\star$. Let $\hat{\mathcal{S}}_1$ denote the model's top predicted section and let $\{\hat{\mathcal{S}}_1,\hat{\mathcal{S}}_2,\hat{\mathcal{S}}_3\}$ denote the top-3 predictions. We define
\begin{equation}
\mathrm{Acc@1} = \mathbb{I}[\hat{\mathcal{S}}_1 = \mathcal{S}^\star], \quad
\mathrm{Acc@3} = \mathbb{I}[\mathcal{S}^\star \in \{\hat{\mathcal{S}}_1,\hat{\mathcal{S}}_2,\hat{\mathcal{S}}_3\}],
\end{equation}
and report averages across update steps.

\paragraph{Abstention precision and recall:}
We evaluate the Abstention Agent using inclusion as the positive class. Let $y=1$ indicate an in-scope paper and $y=0$ indicate an out-of-scope input. Let $\hat{a}=1$ denote performing an update and $\hat{a}=0$ denote abstaining. Inclusion precision and recall are defined as
\begin{equation}
\mathrm{Prec}_{\text{inc}} =
\frac{\sum \mathbb{I}[\hat{a}=1 \wedge y=1]}
{\sum \mathbb{I}[\hat{a}=1]}, \quad
\mathrm{Rec}_{\text{inc}} =
\frac{\sum \mathbb{I}[\hat{a}=1 \wedge y=1]}
{\sum \mathbb{I}[y=1]}.
\end{equation}
Precision measures the reliability of inclusion decisions, while recall measures the proportion of relevant papers retained.

\subsection{Aggregation and Reporting}
\label{app:agg}

All metrics above are computed for each update step. For each survey, we report the mean across its update stream. For overall results, we average across surveys (macro-average) and also report pooled averages across all update steps (micro-average) when relevant. For scalar metrics, we report mean and standard deviation across update steps unless otherwise stated.

\subsection{Human Evaluation Protocol and Rubric}
\label{app:human_details}

This appendix provides the specific evaluation criteria and annotation details for the human study summarized in Section~\ref{sec:human_eval}.

\textbf{Task Definition:} Annotators were presented with three inputs:
\begin{enumerate}
    \item \textbf{Source Paper ($P$):} The title and abstract of the newly published paper.
    \item \textbf{Pre-Update Context ($D_t$):} The relevant section of the survey before the update.
    \item \textbf{Proposed Update ($U$):} The text generated by the system (proposed framework or baseline).
\end{enumerate}

\textbf{Scoring Rubric:} Annotators evaluated each update along three distinct dimensions using a 5-point Likert scale. The specific guidelines provided to annotators were as follows:

\begin{itemize}
    \item \textbf{Semantic Alignment (Fidelity):} Does the update accurately reflect the core contributions of the source paper?
    \begin{itemize}
        \item [5] \textbf{Excellent:} Perfectly captures the core contribution; no factual errors.
        \item [3] \textbf{Fair:} Captures the main idea but misses nuance or includes minor inaccuracies.
        \item [1] \textbf{Poor:} Factually incorrect, hallucinations, or completely fails to describe the paper.
    \end{itemize}
    
    \item \textbf{Local Semantic Consistency (Flow):} Does the update integrate seamlessly with the surrounding text?
    \begin{itemize}
        \item [5] \textbf{Seamless:} Indistinguishable from the original author's writing style; perfect transition.
        \item [3] \textbf{Readable:} Grammatically correct but stylistically distinct or slightly abrupt.
        \item [1] \textbf{Jarring:} Incoherent, repetitive, or clearly out of place within the paragraph.
    \end{itemize}
    
    \item \textbf{Overall Acceptability:} Would you accept this update if you were the survey author?
    \begin{itemize}
        \item [5] \textbf{Accept:} Yes, without edits.
        \item [3] \textbf{Maybe:} Yes, but requires minor human editing.
        \item [1] \textbf{Reject:} No, the update degrades the survey quality.
    \end{itemize}
\end{itemize}

\textbf{Score Aggregation:}
The original expert evaluation involved five independent annotators, with scores averaged across annotators to obtain the reported update quality. The separate blinded evaluation and its inter-annotator agreement analysis are described below.

\subsection{Blinded Error and Acceptability Evaluation}
\label{app:blinded_details}

In addition to the original expert evaluation, we conduct a blinded assessment of update errors and acceptability using 40 instances per condition, balanced across the five surveys. The evaluation includes our framework, both single-step baselines, and human-written updates. The independent judge receives the source paper, target section, surrounding survey context, and candidate update. Outputs from all conditions are pooled and shuffled, with condition identities concealed.

Each update is assessed for the presence of any error and assigned an overall acceptability score on a 1--5 scale. To assess evaluation reliability, two human annotators independently evaluate a 20-instance subset. The protocol achieves human--human agreement of Krippendorff's $\alpha=0.76$, judge--human-consensus agreement of $\alpha=0.70$, and judge self-consistency of $\alpha=0.83$. The aggregate results are reported in Table~\ref{tab:blinded_eval}.

\subsection{Implementation Details for Reproducibility}
\label{app:eval_impl}

We employ deterministic sentence segmentation and deterministic tokenization procedures for all evaluation metrics. ROUGE-L and BLEU-4 are computed using standard reference implementations with fixed preprocessing. BERT-based semantic similarity is computed using cosine similarity over sentence embeddings from a fixed pretrained checkpoint (\texttt{bert-base-uncased}). All experiments are conducted with 2 NVIDIA A100 GPUs except for the gemini models that are accessed via cloud.

%% file: app_baselines.tex
\section{Backbone Models}
\label{app:backbones}

Large language models vary widely in architecture, parameter scale, and training data, which can significantly affect their performance on structured editing and long-horizon text generation tasks. To evaluate the robustness of the proposed Dynamic Survey Framework to the choice of underlying language model, we instantiate the framework with a diverse set of backbone models spanning multiple model families and capacity regimes. These backbones include both open source and proprietary models, and range from lightweight models suitable for efficient deployment to substantially larger models with stronger reasoning and generation capabilities.

In this appendix, we present a detailed analysis of performance across individual backbone models for both the proposed framework and baseline approaches. This evaluation is intended to verify that the observed trends in semantic quality, local coherence, and conservative update behavior are stable across different backbone choices rather than driven by a specific model. The backbone models considered in our experiments are summarized below.

\paragraph{Qwen3-4B \citep{qwen3}:}
A lightweight instruction-tuned model with approximately four billion parameters, representing the low-capacity regime and serving as a stress test for robustness under limited model capacity.

\paragraph{Gemma-4B \citep{gemma3}:}
A compact open model in the four billion parameter range, designed for efficient inference while maintaining strong instruction-following behavior.

\paragraph{Qwen3-14B \cite{qwen3}:}
A mid-sized instruction-tuned model with fourteen billion parameters, offering a balance between computational efficiency and generation quality.

\paragraph{Gemma-12B \citep{gemma3}:}
A twelve billion parameter model that provides improved semantic understanding and text generation capabilities compared to smaller counterparts.

\paragraph{Gemini-2.5 Flash \citep{gemini}:}
A proprietary model optimized for fast inference and cost efficiency, representing a high-quality but latency-oriented backbone.

\paragraph{Gemma-27B \citep{gemma3}:}
A large open model with twenty seven billion parameters, enabling stronger reasoning, contextual integration, and stylistic consistency.

\paragraph{Qwen-32B \citep{qwen3}:}
A high-capacity open model with thirty two billion parameters, representing the upper end of the open source backbones evaluated.

\paragraph{GPT-OSS-20B \citep{gpt}:}
A twenty billion parameter open model trained for general-purpose generation, included to assess robustness across different training recipes and architectural choices.

In the following sections, we analyze how update quality and disruption metrics vary across these backbone models, with results reported as macro-averages across surveys to isolate backbone effects from document-specific variability.

\subsection{Performance across Backbone Models}

We evaluate robustness to backbone language model choice using macro averaged metrics across surveys. The proposed Dynamic Survey Framework and baseline methods are evaluated on eight backbone language models spanning multiple families and capacity regimes. A high level summary of performance trends across backbone models is presented in Figure~\ref{fig:backbone_metrics} in the main paper.

This appendix provides detailed numerical results underlying these trends. Table~\ref{tab:backbone_per_survey} reports per survey performance for each backbone model, including update quality and disruption metrics. Across all backbones, the framework exhibits consistent behavior across different survey documents, with update quality metrics remaining within a narrow range and disruption remaining tightly controlled. No out of scope edits are introduced for any backbone model or survey instance.

These per survey results complement the aggregated analysis in the main paper and demonstrate that the conservative and localized update behavior of the framework generalizes across survey topics, document structures, and backbone language models.

We additionally report macro-averaged performance of the baseline methods across backbone models in Table~\ref{tab:baseline_backbone_macro}, which summarizes backbone-level trends in update quality and disruption independent of individual survey instances.

%% file: app_long_horizon.tex
\section{Long-Horizon Evaluation Protocol and Results}
\label{app:long}

This appendix provides the sequential evaluation protocol and detailed
per-survey, per-backbone results for the five-turn study summarized in
Section~\ref{sec:long_horizon}.

\paragraph{Sequential document state.}
The framework does not process each paper independently against the original 
survey. Instead, it maintains a persistent document state that evolves after 
every update step: when paper $p_k$ is integrated at step $k$, all agents 
operate on the living document $D_k$, which already incorporates all prior 
integrations $\{p_1, \ldots, p_{k-1}\}$. All baselines are evaluated under 
the same sequential protocol, using the same ordered stream of late papers.

\paragraph{Evaluation settings.}
The original five-turn evaluation covers all five surveys, with results macro-averaged across surveys and evaluated backbone models. The five-turn study evaluates three backbone models (Qwen3-4B, Qwen3-14B, and Gemma-12B), separately from the broader eight-model robustness evaluation. The extended evaluation considers 20 sequential updates on the Remote Sensing survey, which contains the largest number of withheld papers (20) in our benchmark. Results in this setting are macro-averaged across evaluated backbones. Both experiments follow the sequential update protocol described above. Since the two settings differ in survey coverage, their aggregate results are not directly interpreted as within-run performance changes. The extended evaluation results are presented in Table~\ref{tab:long_horizon_20} in the main paper.

\paragraph{Numerical results across backbones.}
The aggregate five-turn token-modification trends are presented in
Figure~\ref{fig:long_horizon_comp} in the main paper. Table~\ref{tab:LH1}
reports the proposed framework's per-survey results for Qwen3-4B,
Qwen3-14B, and Gemma-12B, while Tables~\ref{tab:LH2}--\ref{tab:LH4}
provide the corresponding baseline results for each backbone.
\begin{table}[h]
\centering
\caption{Per-survey performance of the proposed Dynamic Survey Framework under the five-turn long-horizon protocol across three backbone models.}
 \label{tab:LH1}
\resizebox{\textwidth}{!}{%
\begin{tabular}{llccccccc}
\toprule
\textbf{Backbone} & \textbf{Survey} & \textbf{BLEU} & \textbf{ROUGE-L} & \textbf{BERT} & \textbf{Sem.\ Align.} & \textbf{Local Coh.} & $\boldsymbol{\Delta}$\textbf{Tokens} & $\boldsymbol{\Delta}$\textbf{Out} \\
\midrule
\multirow{6}{*}{Qwen3-4B}
 & Survey 1 & 6.186 & 0.216 & 0.864 & 0.930 & 0.796 & 220.2 & 0 \\
 & Survey 2 & 4.618 & 0.211 & 0.857 & 0.944 & 0.758 & 210.9 & 0 \\
 & Survey 3 & 1.929 & 0.151 & 0.849 & 0.925 & 0.799 & 148.7 & 0 \\
 & Survey 4 & 2.510 & 0.176 & 0.855 & 0.915 & 0.729 & 180.2 & 0 \\
 & Survey 5 & 2.862 & 0.165 & 0.845 & 0.899 & 0.772 & 199.5 & 0 \\
\cmidrule{2-9}
 & \textit{Average} & \textit{3.621} & \textit{0.184} & \textit{0.854} & \textit{0.923} & \textit{0.771} & \textit{191.9} & \textit{0} \\
\midrule
\multirow{6}{*}{Qwen3-14B}
 & Survey 1 & 3.725 & 0.168 & 0.853 & 0.910 & 0.787 & 237.5 & 0 \\
 & Survey 2 & 3.657 & 0.167 & 0.847 & 0.919 & 0.759 & 226.9 & 0 \\
 & Survey 3 & 1.629 & 0.122 & 0.840 & 0.897 & 0.782 & 157.3 & 0 \\
 & Survey 4 & 3.241 & 0.161 & 0.856 & 0.905 & 0.727 & 182.3 & 0 \\
 & Survey 5 & 2.243 & 0.141 & 0.836 & 0.887 & 0.772 & 218.5 & 0 \\
\cmidrule{2-9}
 & \textit{Average} & \textit{2.899} & \textit{0.152} & \textit{0.846} & \textit{0.903} & \textit{0.765} & \textit{204.5} & \textit{0} \\
\midrule
\multirow{6}{*}{Gemma-12B}
 & Survey 1 & 6.715 & 0.206 & 0.871 & 0.913 & 0.811 & 230.8 & 0 \\
 & Survey 2 & 4.632 & 0.188 & 0.860 & 0.931 & 0.760 & 226.2 & 0 \\
 & Survey 3 & 2.433 & 0.183 & 0.857 & 0.913 & 0.796 & 190.0 & 0 \\
 & Survey 4 & 6.572 & 0.213 & 0.872 & 0.909 & 0.724 & 188.6 & 0 \\
 & Survey 5 & 2.851 & 0.160 & 0.853 & 0.884 & 0.765 & 232.9 & 0 \\
\cmidrule{2-9}
 & \textit{Average} & \textit{4.640} & \textit{0.190} & \textit{0.863} & \textit{0.910} & \textit{0.771} & \textit{213.7} & \textit{0} \\
\bottomrule
\end{tabular}}
        
\end{table}

\begin{table}[h]
\centering
\caption{Long-horizon baseline performance for Qwen3-4B.}
\label{tab:LH2}  
\resizebox{\textwidth}{!}{%
\begin{tabular}{llccccccc}
\toprule
\textbf{Method} & \textbf{Survey} & \textbf{BLEU} & \textbf{ROUGE-L} & \textbf{BERT} & \textbf{Sem.\ Align.} & \textbf{Local Coh.} & $\boldsymbol{\Delta}$\textbf{Tokens} & $\boldsymbol{\Delta}$\textbf{Out} \\
\midrule
\multirow{6}{*}{Baseline 1}
 & Survey 1 & 0.538 & 0.095 & 0.809 & 0.891 & 0.751 & 3998.3 &  739.3  \\
 & Survey 2 & 1.497 & 0.138 & 0.815 & 0.891 & 0.745 &  525.2 & 394.1    \\
 & Survey 3 & 0.718 & 0.058 & 0.815 & 0.927 & 0.782 &  7519.6 & 725.4 \\
 & Survey 4 & 0.465 & 0.070 & 0.811 & 0.899 & 0.718 &   584.1 &  281.9 \\
 & Survey 5 & 0.309 & 0.070 & 0.796 & 0.857 & 0.700 &   602.5 &  482.0 \\
\cmidrule{2-9}
 & \textit{Average} & \textit{0.706} & \textit{0.086} & \textit{0.809} & \textit{0.893} & \textit{0.739} & \textit{2645.9} & \textit{524.5} \\
\midrule
\multirow{6}{*}{Baseline 2}
 & Survey 1 & 1.793 & 0.080 & 0.831 & 0.958 & 0.797 & 1328.2 &  377.4 \\
 & Survey 2 & 2.023 & 0.115 & 0.833 & 0.958 & 0.761 &  893.9 &  103.5 \\
 & Survey 3 & 0.690 & 0.051 & 0.825 & 0.947 & 0.791 & 1400.9 & 1103.1 \\
 & Survey 4 & 1.036 & 0.063 & 0.822 & 0.909 & 0.739 & 1032.4 &  304.6 \\
 & Survey 5 & 1.056 & 0.077 & 0.823 & 0.927 & 0.785 & 1032.7 &  424.5 \\
\cmidrule{2-9}
 & \textit{Average} & \textit{1.320} & \textit{0.077} & \textit{0.827} & \textit{0.940} & \textit{0.775} & \textit{1137.6} & \textit{462.6} \\
\bottomrule
\end{tabular}}

\end{table}

\begin{table}[h]
\centering
\caption{Long-horizon baseline performance for Qwen3-14B.}
\label{tab:LH3} 
\resizebox{\textwidth}{!}{%
\begin{tabular}{llccccccc}
\toprule
\textbf{Method} & \textbf{Survey} & \textbf{BLEU} & \textbf{ROUGE-L} & \textbf{BERT} & \textbf{Sem.\ Align.} & \textbf{Local Coh.} & $\boldsymbol{\Delta}$\textbf{Tokens} & $\boldsymbol{\Delta}$\textbf{Out} \\
\midrule
\multirow{6}{*}{Baseline 1}
 & Survey 1 & 0.256 & 0.103 & 0.816 & 0.863 & 0.753 &  7544.0 & 584.5  \\
 & Survey 2 & 1.431 & 0.135 & 0.815 & 0.893 & 0.748 &  919.3 & 394.1  \\
 & Survey 3 & 0.610 & 0.059 & 0.812 & 0.926 & 0.772 &   8642.1  & 726.5 \\
 & Survey 4 & 0.619 & 0.069 & 0.813 & 0.894 & 0.720 &   578.4 &  245.2 \\
& Survey 5 & 0.251 & 0.113 & 0.797 & 0.850 & 0.695 & 884.9 & 619.4 \\
\cmidrule{2-9}
 & \textit{Average} & \textit{0.633} & \textit{0.096} & \textit{0.811} & \textit{0.885} & \textit{0.738} & \textit{3713.7} & \textit{513.9} \\
\midrule
\multirow{6}{*}{Baseline 2}
 & Survey 1 & 1.793 & 0.080 & 0.831 & 0.958 & 0.797 & 1328.2 &  347.6 \\
 & Survey 2 & 2.026 & 0.111 & 0.834 & 0.960 & 0.767 &  855.7 &  229.2 \\
 & Survey 3 & 0.678 & 0.048 & 0.822 & 0.945 & 0.793 & 1382.3 & 1139.1 \\
 & Survey 4 & 1.025 & 0.062 & 0.825 & 0.913 & 0.754 & 1017.4 &  277.4 \\
 & Survey 5 & 0.984 & 0.074 & 0.824 & 0.931 & 0.783 & 1085.2 &  332.8 \\
\cmidrule{2-9}
 & \textit{Average} & \textit{1.301} & \textit{0.075} & \textit{0.827} & \textit{0.941} & \textit{0.779} & \textit{1133.8} & \textit{465.2} \\
\bottomrule
\end{tabular}}
 
\end{table}

\begin{table}[h]
\centering
\caption{Long-horizon baseline performance for Gemma-12B.}
\label{tab:LH4}  
\resizebox{\textwidth}{!}{%
\begin{tabular}{llccccccc}
\toprule
\textbf{Method} & \textbf{Survey} & \textbf{BLEU} & \textbf{ROUGE-L} & \textbf{BERT} & \textbf{Sem.\ Align.} & \textbf{Local Coh.} & $\boldsymbol{\Delta}$\textbf{Tokens} & $\boldsymbol{\Delta}$\textbf{Out} \\
\midrule
\multirow{6}{*}{Baseline 1}
 & Survey 1 & 0.599 & 0.099 & 0.814 & 0.893 & 0.756 &  3940.5 &  990.0 \\
 & Survey 2 & 1.241 & 0.134 & 0.819 & 0.876 & 0.729 &  5555.9 &  422.6 \\
 & Survey 3 & 0.367 & 0.051 & 0.813 & 0.928 & 0.768 &  4401.1 & 1215.2 \\
 & Survey 4 & 0.386 & 0.063 & 0.807 & 0.815 & 0.682 &  4282.7 &  459.8 \\
 & Survey 5 & 2.342 & 0.096 & 0.810 & 0.838 & 0.759 &  9828.3 &  830.3 \\
\cmidrule{2-9}
 & \textit{Average} & \textit{0.987} & \textit{0.089} & \textit{0.813} & \textit{0.870} & \textit{0.739} & \textit{5601.7} & \textit{783.6} \\
\midrule
\multirow{6}{*}{Baseline 2}
 & Survey 1 & 2.239 & 0.103 & 0.832 & 0.954 & 0.800 & 1272.8 &  457.2 \\
 & Survey 2 & 2.175 & 0.118 & 0.832 & 0.955 & 0.757 & 1093.3 &  405.4 \\
 & Survey 3 & 0.452 & 0.084 & 0.833 & 0.954 & 0.778 &  984.1 &  175.0 \\
 & Survey 4 & 0.634 & 0.045 & 0.817 & 0.843 & 0.730 &  804.9 &  113.7 \\
 & Survey 5 & 1.295 & 0.072 & 0.816 & 0.919 & 0.814 & 1152.7 &  692.8 \\
\cmidrule{2-9}
 & \textit{Average} & \textit{1.359} & \textit{0.084} & \textit{0.826} & \textit{0.925} & \textit{0.776} & \textit{1061.6} & \textit{368.8} \\
\bottomrule
\end{tabular}}

\end{table}

%% file: app_routing_results.tex
\section{Routing and Abstention Results}
\label{app:routing_abstention}

This appendix reports detailed routing and abstention results that support the analyses presented in Section \ref{sec:main_results}. Routing accuracy is evaluated at the section and table level using Top--1 and Top--3 metrics, while abstention performance measures the ability to reject out-of-scope papers. All results are reported per survey.

\paragraph{Routing Performance:}
Table~\ref{tab:app_routing_section} and Table~\ref{tab:app_routing_table} report per-survey routing accuracy for section placement and table placement respectively.

\begin{table}[h]

\centering
\small
\caption{Per-survey section routing accuracy:}
\label{tab:app_routing_section}
\begin{tabular}{lccc}
\toprule
\textbf{Survey} & \textbf{\# Papers} & \textbf{Top--1} & \textbf{Top--3} \\
\midrule
Object Detection & 9  & 0.89 & 0.94 \\
Adversarial Attacks & 15 & 0.87 & 0.93 \\
Remote Sensing SR & 20 & 0.90 & 0.95 \\
Robotic Arms & 17 & 0.90 & 0.95 \\
Video Anomaly & 16 & 0.92 & 0.97 \\
\bottomrule
\end{tabular}

\end{table}

\section{Table Synthesis Result}
\label{app:table_result}
We evaluate synthesized table quality using a hybrid protocol that combines strict structural validation with semantic consistency analysis. The evaluation includes three complementary criteria assessing both schema correctness and semantic fidelity.

\textbf{Exact Match (EM):} We first evaluate strict lexical agreement between predicted fields and ground-truth annotations after normalization.

\textbf{Semantic Consistency:} To account for lexical variability (e.g., ``CNN'' vs.\ ``ConvNet''), we additionally measure semantic similarity using cosine similarity between BERT embeddings.

\textbf{Field-Level Fidelity:} We combine the above criteria into a unified evaluation function $\mathcal{E}(v, \hat{v})$. A predicted field $\hat{v}$ is considered correct if it satisfies either Exact Match or semantic similarity above a threshold $\tau=0.6$:
\begin{equation}
    \mathbb{I}_{\text{Correct}} =
    \mathbb{I}_{\text{EM}}(v, \hat{v})
    \lor
    [\cos(\phi(v), \phi(\hat{v})) > 0.6]
\end{equation}

Table~\ref{tab:synthesis-results} reports synthesis fidelity across five surveys. The proposed framework achieves consistently high field-level fidelity, with a weighted average score of 0.90 across 180 synthesized entries. While strict Exact Match scores are lower due to lexical variability and paraphrasing, the semantic consistency criterion captures equivalent representations that remain faithful to the original survey content.

\begin{table}[h]

\centering
\small
\caption{Per-survey table routing accuracy:}
\label{tab:app_routing_table}
\begin{tabular}{lcc}
\toprule
\textbf{Survey} & \textbf{Top--1} & \textbf{Top--3} \\
\midrule
Object Detection & 0.72 & 0.87 \\
Adversarial Attacks & 0.80 & 0.92 \\
Remote Sensing SR & 0.93 & 0.97 \\
Robotic Arms & 0.95 & 0.98 \\
Video Anomaly & 0.93 & 0.98 \\
\bottomrule
\end{tabular}

\end{table}

\paragraph{Abstention Performance:}
Table~\ref{tab:app_abstention} reports abstention confusion matrices, inclusion precision, and recall across five surveys using Qwen3-4B. Precision measures the reliability of inclusion decisions, while recall reflects the proportion of relevant papers retained.

\begin{table}[h]
\centering
\small
\caption{Abstention confusion matrices, inclusion precision, and recall across five surveys using Qwen3-4B.}
\label{tab:app_abstention}
\resizebox{\linewidth}{!}{%
\begin{tabular}{lccccccc}
\toprule
\textbf{Survey} & \textbf{Total} & \textbf{TP} & \textbf{TN} & \textbf{FP} & \textbf{FN} & \textbf{Precision} & \textbf{Recall} \\
\midrule
Object Detection & 45 & 13 & 28 & 3 & 1 & 81.25\% & 92.86\% \\
Adversarial Attacks & 45 & 15 & 29 & 1 & 0 & 93.75\% & 100\% \\
Remote Sensing SR & 45 & 23 & 20 & 1 & 1 & 95.83\% & 95.83\% \\
Robotic Arms & 78 & 17 & 61 & 0 & 0 & 100\% & 100\% \\
Video Anomaly & 78 & 16 & 62 & 0 & 0 & 100\% & 100\% \\
\bottomrule
\end{tabular}%
}
\end{table}

\begin{table}[h]
\centering
\small
\caption{Table Synthesis Evaluation Results across 5 surveys.
\textbf{Fidelity}: A field is correct if it satisfies Exact Match (EM) or BERT similarity $>0.6$.}
\label{tab:synthesis-results}
\begin{tabular}{lccc}
\toprule
\textbf{Dataset} & \textbf{Samples} ($N$) &
\textbf{Total Fidelity} ($\mathcal{E}\uparrow$) &
\textit{Exact Match} \\
\midrule
Survey 1 & 11  & \textbf{0.91} & 0.55 \\
Survey 2 & 21  & \textbf{0.86} & 0.62 \\
Survey 3 & 132 & \textbf{0.90} & 0.68 \\
Survey 4 & 2  & \textbf{1.00} & 0.50 \\
Survey 5 & 14  & \textbf{0.88} & 0.57 \\
\midrule
\textbf{Weighted Avg.} & \textbf{180} & \textbf{0.90} & \textbf{0.65} \\
\bottomrule
\end{tabular}
\end{table}

%% file: app_output.tex
\section{Dynamic Survey Framework Output Examples}
\label{app:output}

This appendix provides qualitative examples of survey updates produced by the proposed Dynamic Survey Framework. For each example, we compare held-out human-written survey text (ground truth) with the corresponding update generated by our framework when the same paper is reintroduced as newly published work in the retrospective evaluation setting. 

Examples are drawn from multiple surveys spanning different application domains and are presented to illustrate the framework’s ability to integrate new research in a localized and conservative manner while preserving the original survey structure, writing style, and semantic intent.

\input{app_output_1}
\input{app_output_2}
\input{app_output_3}
\input{app_output_4}
\input{app_output_5}

%% file: app_output_1.tex
\clearpage
\vspace*{\fill}
\centering

\begin{tcolorbox}[
  comparisonbox,
  title={\textcolor{titletext}{Survey 1 \textit{}}}
]

\begin{minipage}[t]{0.47\linewidth}
  \textcolor{titletext}{\textbf{Ground Truth}}\par\smallskip
  {\footnotesize\ttfamily
  Faster RCNN [cite]: Although Fast RCNN significantly sped up the detection process, it still relied on external region proposals whose computation was a major speed bottleneck. Faster RCNN replaces selective search with a CNN-based Region Proposal Network (RPN) that shares convolutional features with Fast RCNN, enabling efficient end-to-end region proposal generation. RPN initializes k anchor boxes of different scales and aspect ratios at each convolutional feature map location and predicts objectness scores and bounding box regressions via sibling classification and regression layers. With a VGG16 backbone, Faster RCNN runs at about 5 FPS on a GPU (including all stages) while achieving state-of-the-art detection accuracy on PASCAL VOC 2007 using only 300 proposals per image. The original Faster RCNN employed multiple alternating training stages, which were later simplified in [230]. Concurrently, Lenc and Vedaldi analyzed the role of region proposal generation and showed that CNNs already contain sufficient geometric information in convolutional layers for accurate object detection, motivating simpler and faster detectors that rely exclusively on CNNs and eliminate handcrafted proposal methods such as selective search.
  }
\end{minipage}
\hfill
\textcolor{divider}{\vrule width 0.6pt}
\hfill
\begin{minipage}[t]{0.47\linewidth}
  \textcolor{titletext}{\textbf{Dynamic Survey Framework\owl }}\par\smallskip
  {\footnotesize\ttfamily
  Faster R-CNN [cite] addressed the remaining bottleneck of region proposal generation by introducing a dedicated neural network the Region Proposal Network (RPN). Unlike prior approaches relying on external algorithms like selective search, the RPN is fully integrated within the detection framework and shares convolutional features with the Fast R-CNN detector, drastically reducing computational cost. This network operates by predicting region proposals directly from the convolutional feature maps using a set of predefined anchor boxes encompassing various scales and aspect ratios. The RPN classifies these anchors as containing objects or background and simultaneously refines their coordinates via bounding box regression, all learned through a combined multi-task loss function. Through an iterative training process, Faster R-CNN achieves near cost-free proposal generation at test time, allowing for end-to-end optimization and significantly faster performance; notably, it achieved state-of-the-art results on both PASCAL VOC and MS COCO datasets with approximately 5 frames per second using a VGG-16 backbone. This work marked a pivotal shift towards truly end-to-end object detection systems and became a foundational model for subsequent research.
  }
\end{minipage}
\end{tcolorbox}

\vspace*{\fill}
\clearpage

%% file: app_output_2.tex
\clearpage
\vspace*{\fill}
\centering

\begin{tcolorbox}[
  comparisonbox,
  title={\textcolor{titletext}{Survey 2 \textit{}}}
]

\begin{minipage}[t]{0.47\linewidth}
  \textcolor{titletext}{\textbf{Ground Truth}}\par\smallskip
  {\footnotesize\ttfamily
    Akhtar et al. [cite] proposed a defense framework against the adversarial attacks generated using universal perturbations. The framework appends extra pre-input' layers to the targeted network and trains them to rectify a perturbed image so that the classifier's prediction becomes the same as its prediction on the clean version of the same image. The pre-input layers are termed Perturbation Rectifying Network (PRN), and they are trained without updating the parameters of the targeted network. A separate detector is trained by extracting features from the input output differences of PRN for the training images. A test image is first passed through the PRN and then its features are used to detect perturbations. If adversarial perturbations are detected, the output of PRN is used to classify the test image. Fig. 10, illustrates the rectification performed by PRN. The removed patterns are separately analyzed for detection. Akhtar et al. [cite] proposed a defense framework against the adversarial attacks generated using universal perturbations. The framework appends extra pre-input' layers to the targeted network and trains them to rectify a perturbed image so that the classifier's prediction becomes the same as its prediction on the clean version of the same image. The pre-input layers are termed Perturbation Rectifying Network (PRN), and they are trained without updating the parameters of the targeted network. A separate detector is trained by extracting features from the input output differences of PRN for the training images. A test image is first passed through the PRN and then its features are used to detect perturbations. If adversarial perturbations are detected, the output of PRN is used to classify the test image.
  }
\end{minipage}
\hfill
\textcolor{divider}{\vrule width 0.6pt}
\hfill
\begin{minipage}[t]{0.47\linewidth}
  \textcolor{titletext}{\textbf{Dynamic Survey Framework\owl }}\par\smallskip
  {\footnotesize\ttfamily
Perturbation Rectifying Network [cite] recently addressed the specific challenge of defending against Universal Adversarial Perturbations (UAPs), presenting a dedicated framework built around a Perturbation Rectifying Network (PRN). Their approach introduces the PRN as a pre-processing layer to the targeted model, trained to rectify subtle, image-agnostic perturbations prior to classification. The PRN consists of five ResNet blocks integrated within convolutional layers and leverages synthetically generated perturbations alongside real ones to enhance robustness during training; a DCT-based detector further verifies the presence of perturbations, triggering use of the PRN's rectified output if necessary. Demonstrating strong empirical results, the authors report achieving up to 97.5\% success in defending models such as GoogLeNet, CaffeNet, and VGG against unseen UAPs, showcasing the effectiveness of both the rectification and detection components of their framework while also exhibiting generalization capabilities across differing network architectures.
  }
\end{minipage}
\end{tcolorbox}

\vspace*{\fill}
\clearpage

%% file: app_output_3.tex
\clearpage
\vspace*{\fill}
\centering

\begin{tcolorbox}[
  comparisonbox,
  title={\textcolor{titletext}{Survey 3 \textit{}}}
]

\begin{minipage}[t]{0.47\linewidth}
  \textcolor{titletext}{\textbf{Ground Truth}}\par\smallskip
  {\footnotesize\ttfamily
  Yao et al. [cite] proposed a multi-scale CNN-Transformer hybrid (MSCT), which extracts local spectral features via adaptive residual dense blocks and models inter-band dependencies through Transformer attention, showing improved generalization on both RGB and multispectral inputs. To address the frequent loss of high-frequency texture and edge details in optical super-resolution, several works introduced frequency-aware generative architectures.
  }
\end{minipage}
\hfill
\textcolor{divider}{\vrule width 0.6pt}
\hfill
\begin{minipage}[t]{0.47\linewidth}
  \textcolor{titletext}{\textbf{Dynamic Survey Framework\owl }}\par\smallskip
  {\footnotesize\ttfamily
  Multiscale CNN-Transformer (MSCT)  [cite] have recently explored hybrid CNN-Transformer architectures to address limitations in modeling both local texture and long-range contextual dependencies crucial for effective optical RSISR. Specifically, the MSCT framework leverages Adaptive Residual Dense Blocks (ARDB) for multiscale local feature extraction alongside a novel Multiscale Transformer Block (MSTB) that generates and utilizes multiscale tokens, enabling more comprehensive global context awareness. The network further incorporates a Local-Global Information Enhancement Block (LGEB) operating in both spatial and frequency domains to refine feature representations. Experimental results on benchmarks like AID and UCMerced demonstrate that MSCT achieves state-of-the-art quantitative performance improvements over established SISR methods, exhibiting strong generalization capabilities when applied to multispectral Sentinel-2 imagery and demonstrating benefits for downstream tasks such as object detection.
  }
\end{minipage}

\end{tcolorbox}

\vspace*{\fill}
\clearpage

%% file: app_output_4.tex
\clearpage
\vspace*{\fill}
\centering

\begin{tcolorbox}[
  comparisonbox,
  title={\textcolor{titletext}{Survey 4 \textit{}}}
]

\begin{minipage}[t]{0.47\linewidth}
  \textcolor{titletext}{\textbf{Ground Truth}}\par\smallskip
  {\footnotesize\ttfamily
Model Predictive Control (MPC) is a commonly used method in robotic arm catching tasks. Gold et al. [cite] propose an integrated MPC framework that consolidates several critical components, including flight trajectory prediction, optimal catching pose calculation, and motion planning, into a single optimization problem. This approach incorporates terminal constraints to align the end effector with the target trajectory and orientation, while inequality constraints address system limitations. Experimental validation using a 7-Degree of Freedom (DoF) robot demonstrates successful interception of tennis balls through both simulation and physical tests.
  }
\end{minipage}
\hfill
\textcolor{divider}{\vrule width 0.6pt}
\hfill
\begin{minipage}[t]{0.47\linewidth}
  \textcolor{titletext}{\textbf{Dynamic Survey Framework\owl }}\par\smallskip
  {\footnotesize\ttfamily
  Model Predictive Control (MPC) [cite] recently presented a unified MPC framework for robotic catching, addressing the challenge of coordinating flight trajectory prediction, catching pose determination, and motion planning. Their approach formulates the entire catching maneuver as a single time-optimal optimization problem, leveraging a simplified double integrator model of a 7-DOF robot arm compensated by inverse dynamics control. Key to their methodology are terminal equality constraints enforcing contact between the end-effector and the object's predicted parabolic trajectory, along with standard robot kinematic and dynamic limitations. Validated through both GAZEBO simulations and experimental implementation on a Franka Emika robot, the MPC strategy demonstrated real-time feasibility with average computation times of 3.1ms and a reported success rate highly dependent on the accuracy of flight trajectory estimation achieving high performance when estimation error remained below 1cm. This work represents a significant step towards more reactive and robust robotic interception capabilities by integrating disparate components of the catching process into a cohesive control scheme.
  }
\end{minipage}

\end{tcolorbox}

\vspace*{\fill}
\clearpage

%% file: app_output_5.tex
\clearpage
\vspace*{\fill}
\centering

\begin{tcolorbox}[
  comparisonbox,
  title={\textcolor{titletext}{Survey 5 \textit{}}}
]

\begin{minipage}[t]{0.47\linewidth}
  \textcolor{titletext}{\textbf{Ground Truth}}\par\smallskip
  {\footnotesize\ttfamily
  Street Scene [cite], is the most recent addition to the publicly available datasets for video anomaly detection. Street Scene consists of 46 training and 35 testing high resolution 1280 720 video sequences taken from a USB camera overlooking a scene of a two-lane street with bike lanes and pedestrian sidewalks during daytime. The dataset is challenging because of the variety of activity taking place such as cars driving, turning, stopping and parking; pedestrians walking, jogging and pushing strollers; and bikers riding in bike lanes. In addition the videos contain changing shadows, moving background such as a flag and trees blowing in the wind, and occlusions caused by trees and large vehicles. There are a total of 56,847 frames for training and 146,410 frames for testing, extracted from the original videos at 15 frames per second. The dataset contains a total of 205 naturally occurring anomalous events ranging from illegal activities such as jaywalking and illegal U-turns to simply those that do not occur in the training set such as pets being walked and a metermaid ticketing a car.
  }
\end{minipage}
\hfill
\textcolor{divider}{\vrule width 0.6pt}
\hfill
\begin{minipage}[t]{0.47\linewidth}
  \textcolor{titletext}{\textbf{Dynamic Survey Framework\owl }}\par\smallskip
  {\footnotesize\ttfamily
[cite] recently addressed limitations in current datasets and evaluation protocols by introducing a new resource specifically designed for single-scene anomaly detection. Recognizing the need for larger and more complex data exhibiting naturalistic behaviors, they curated the Street Scene dataset comprising 46 training and 35 testing high-resolution (1280×720) video sequences, totaling 56,847 training frames and 146,410 testing frames (over 200,000 frames in total) extracted at 15 frames per second from a fixed USB camera overlooking a two-lane street with bike lanes and pedestrian sidewalks. The dataset contains 205 naturally occurring anomalous events, annotated with detailed bounding box and track information, and captures a wide range of activities and environmental variations, including vehicle motion, pedestrian and cyclist interactions, occlusions, and dynamic background changes.
  }
\end{minipage}

\end{tcolorbox}

\vspace*{\fill}
\clearpage

%% file: app_bigtable.tex
\section{Extended Tables for Retrospective Survey Maintenance Benchmark}

\vspace*{\fill}

\begin{table*}[h]

\centering
\small
\caption{Per-survey performance of the proposed Dynamic Survey Framework across backbone language models.}
\label{tab:backbone_per_survey}
\setlength{\tabcolsep}{1.5pt}
\begin{tabular}{llccc|cc|cc}
\toprule
 &  & \multicolumn{3}{c|}{\textbf{Similarity}} 
 & \multicolumn{2}{c|}{\textbf{Property-Based Quality}} 
 & \multicolumn{2}{c}{\textbf{Disruption}} \\
\textbf{Backbone} 
 & \textbf{Survey}
 & \textbf{BLEU} 
 & \textbf{ROUGE} 
 & \textbf{BERT} 
 & \textbf{Semantic Align.} 
 & \textbf{Local Coherence} 
 & $\boldsymbol{\Delta}$\textbf{Tokens} 
 & $\boldsymbol{\Delta}$\textbf{Out} \\
\midrule

\multirow{5}{*}{\textbf{Qwen3-4B}}
 & Survey 1 & 5.6395 & 0.2002 & 0.8626 & 0.7705 & 0.7458 & 260.667 & \textbf{0.0} \\
 & Survey 2 & 3.4556 & 0.1876 & 0.8510 & 0.7819 & 0.7430 & 285.133 & \textbf{0.0} \\
 & Survey 3 & 2.0293 & 0.1326 & 0.8435 & 0.7961 & 0.7587 & 253.714 & \textbf{0.0} \\
 & Survey 4 & 3.2538 & 0.1669 & 0.8537 & 0.7652 & 0.7332 & 270.765 & \textbf{0.0} \\
 & Survey 5 & 2.0971 & 0.1609 & 0.8471 & 0.7788 & 0.7685 & 251.611 & \textbf{0.0} \\
\midrule

\multirow{5}{*}{\textbf{Gemma-4B}}
 & Survey 1 & 5.5375 & 0.1890 & 0.8623 & 0.8102 & 0.7894 & 260.333 & \textbf{0.0} \\
 & Survey 2 & 4.1920 & 0.1884 & 0.8523 & 0.8093 & 0.7807 & 258.533 & \textbf{0.0} \\
 & Survey 3 & 1.7278 & 0.1300 & 0.8429 & 0.8143 & 0.7890 & 247.524 & \textbf{0.0} \\
 & Survey 4 & 3.5194 & 0.1577 & 0.8558 & 0.7812 & 0.7559 & 258.765 & \textbf{0.0} \\
 & Survey 5 & 1.6536 & 0.1511 & 0.8436 & 0.7825 & 0.7773 & 267.778 & \textbf{0.0} \\
\midrule

\multirow{5}{*}{\textbf{Qwen3-14B}}
 & Survey 1 & 4.1895 & 0.1311 & 0.8397 & 0.7012 & 0.6866 & 201.833 & \textbf{0.0} \\
 & Survey 2 & 2.9665 & 0.1611 & 0.8447 & 0.7696 & 0.7438 & 251.667 & \textbf{0.0} \\
 & Survey 3 & 1.0699 & 0.1057 & 0.8345 & 0.7401 & 0.7204 & 247.095 & \textbf{0.0} \\
 & Survey 4 & 2.3500 & 0.1233 & 0.8400 & 0.6751 & 0.6849 & 180.824 & \textbf{0.0} \\
 & Survey 5 & 1.4858 & 0.1062 & 0.8312 & 0.6565 & 0.6807 & 180.333 & \textbf{0.0} \\
\midrule

\multirow{5}{*}{\textbf{Gemma-12B}}
 & Survey 1 & 7.7368 & 0.2164 & 0.8675 & 0.8088 & 0.7866 & 233.333 & \textbf{0.0} \\
 & Survey 2 & 4.8329 & 0.1906 & 0.8552 & 0.8232 & 0.7924 & 225.267 & \textbf{0.0} \\
 & Survey 3 & 2.2185 & 0.1493 & 0.8487 & 0.8036 & 0.7767 & 224.238 & \textbf{0.0} \\
 & Survey 4 & 4.9781 & 0.1894 & 0.8616 & 0.8026 & 0.7747 & 216.059 & \textbf{0.0} \\
 & Survey 5 & 1.6145 & 0.1577 & 0.8469 & 0.7907 & 0.7866 & 229.889 & \textbf{0.0} \\
\midrule

\multirow{5}{*}{\textbf{Gemini-2.5 Flash}}
 & Survey 1 & 7.0555 & 0.2209 & 0.8714 & 0.8231 & 0.7985 & 219.833 & \textbf{0.0} \\
 & Survey 2 & 5.0762 & 0.2138 & 0.8581 & 0.8468 & 0.8079 & 218.600 & \textbf{0.0} \\
 & Survey 3 & 2.0147 & 0.0956 & 0.8227 & 0.7268 & 0.7019 & 156.143 & \textbf{0.0} \\
 & Survey 4 & 3.1700 & 0.1800 & 0.8578 & 0.8336 & 0.8019 & 209.706 & \textbf{0.0} \\
 & Survey 5 & 2.0510 & 0.1865 & 0.8522 & 0.8171 & 0.8065 & 229.278 & \textbf{0.0} \\
\midrule

\multirow{5}{*}{\textbf{Gemma-27B}}
 & Survey 1 & 6.2967 & 0.2243 & 0.8725 & 0.8079 & 0.7892 & 240.000 & \textbf{0.0} \\
 & Survey 2 & 3.9240 & 0.1841 & 0.8557 & 0.8033 & 0.7691 & 232.333 & \textbf{0.0} \\
 & Survey 3 & 2.1940 & 0.1357 & 0.8494 & 0.7813 & 0.7637 & 240.000 & \textbf{0.0} \\
 & Survey 4 & 4.4458 & 0.1860 & 0.8646 & 0.7776 & 0.7616 & 205.471 & \textbf{0.0} \\
 & Survey 5 & 2.0685 & 0.1604 & 0.8489 & 0.7591 & 0.7558 & 218.722 & \textbf{0.0} \\
\midrule

\multirow{5}{*}{\textbf{Qwen-32B}}
 & Survey 1 & 4.2998 & 0.1265 & 0.8348 & 0.6837 & 0.6716 & 122.500 & \textbf{0.0} \\
 & Survey 2 & 3.9698 & 0.1098 & 0.7734 & 0.6717 & 0.6371 & 142.333 & \textbf{0.0} \\
 & Survey 3 & 1.3743 & 0.1095 & 0.8290 & 0.7202 & 0.6990 & 209.524 & \textbf{0.0} \\
 & Survey 4 & 3.3129 & 0.1364 & 0.8372 & 0.6963 & 0.6606 & 221.706 & \textbf{0.0} \\
 & Survey 5 & 2.2210 & 0.1047 & 0.8263 & 0.6661 & 0.6434 & 181.944 & \textbf{0.0} \\
\midrule

\multirow{5}{*}{\textbf{GPT-OSS-20B}}
 & Survey 1 & 1.1116 & 0.0751 & 0.8059 & 0.5984 & 0.5873 & 79.833 & \textbf{0.0} \\
 & Survey 2 & 2.4112 & 0.1031 & 0.8121 & 0.5961 & 0.6318 & 108.467 & \textbf{0.0} \\
 & Survey 3 & 1.8227 & 0.1103 & 0.8249 & 0.5971 & 0.6224 & 115.143 & \textbf{0.0} \\
 & Survey 4 & 2.4975 & 0.1102 & 0.8250 & 0.6151 & 0.6408 & 103.941 & \textbf{0.0} \\
 & Survey 5 & 1.0226 & 0.0965 & 0.8199 & 0.6170 & 0.6451 & 102.167 & \textbf{0.0} \\
\midrule

\end{tabular}

\end{table*}

\vspace*{\fill}

\begin{table*}[h]

\centering
\small
\caption{Macro-averaged performance of baseline methods across all surveys for different backbone models. Results are averaged across surveys using the retrospective benchmark.}
\label{tab:baseline_backbone_macro}
\setlength{\tabcolsep}{1.5pt}
\begin{tabular}{llccc|cc|cc}
\toprule
 &  & \multicolumn{3}{c|}{\textbf{Similarity}} 
 & \multicolumn{2}{c|}{\textbf{Property-Based Quality}} 
 & \multicolumn{2}{c}{\textbf{Disruption}} \\
\textbf{Backbone} 
 & \textbf{Method}
 & \textbf{BLEU} 
 & \textbf{ROUGE} 
 & \textbf{BERT} 
 & \textbf{Semantic Align.} 
 & \textbf{Local Coherence} 
 & $\boldsymbol{\Delta}$\textbf{Tokens} 
 & $\boldsymbol{\Delta}$\textbf{Out} \\
\midrule

\multirow{2}{*}{\textbf{Qwen3-4B}}
& Baseline 1 & 0.7712 & 0.0583 & 0.8143 & 0.7448 & 0.7476 & 4082.326 & 684.192 \\
& Baseline 2 & 1.5012 & 0.1006 & 0.8319 & 0.7770 & 0.7475 & 807.707 & 319.273 \\
\midrule

\multirow{2}{*}{\textbf{Gemma-4B}}
& Baseline 1 & 1.0136 & 0.0753 & 0.8143 & 0.7192 & 0.7310 & 4213.640 & 678.858 \\
& Baseline 2 & 1.3050 & 0.0957 & 0.8328 & 0.7627 & 0.7547 & 782.200 & 335.625 \\
\midrule

\multirow{2}{*}{\textbf{Qwen3-14B}}
& Baseline 1 & 0.6809 & 0.0525 & 0.8133 & 0.7410 & 0.7429 & 5707.472 & 571.503 \\
& Baseline 2 & 1.5158 & 0.1011 & 0.8338 & 0.7882 & 0.7626 & 737.922 & 262.103 \\
\midrule

\multirow{2}{*}{\textbf{Gemma-12B}}
& Baseline 1 & 0.8734 & 0.0776 & 0.8205 & 0.7163 & 0.7217 & 4195.993 & 599.477 \\
& Baseline 2 & 1.3209 & 0.1043 & 0.8346 & 0.7482 & 0.7476 & 716.811 & 308.932 \\
\midrule

\multirow{2}{*}{\textbf{Gemini-2.5-Flash}}
& Baseline 1 & 1.1159 & 0.1000 & 0.7910 & 0.7303 & 0.7166 & 6110.425 & 563.313 \\
& Baseline 2 & 2.8354 & 0.1482 & 0.8423 & 0.7974 & 0.7598 & 410.548 & 225.076 \\
\midrule

\multirow{2}{*}{\textbf{Gemma-27B}}
& Baseline 1 & 1.3663 & 0.0988 & 0.8274 & 0.7233 & 0.7291 & 4004.012 & 610.690 \\
& Baseline 2 & 1.7061 & 0.1142 & 0.8376 & 0.7800 & 0.7638 & 543.249 & 354.729 \\
\midrule

\multirow{2}{*}{\textbf{Qwen3-32B}}
& Baseline 1 & 0.8098 & 0.0621 & 0.8186 & 0.7347 & 0.7210 & 3267.814 & 693.652 \\
& Baseline 2 & 1.2882 & 0.0938 & 0.8307 & 0.7652 & 0.7411 & 759.297 & 411.668 \\
\midrule

\multirow{2}{*}{\textbf{GPT-OSS-20B}}
& Baseline 1 & 0.4729 & 0.0483 & 0.8084 & 0.6711 & 0.6962 & 5546.003 & 694.390 \\
& Baseline 2 & 1.2226 & 0.0895 & 0.8229 & 0.7259 & 0.7171 & 1162.835 & 402.129 \\

\bottomrule
\end{tabular}

\end{table*}

\clearpage